\documentclass[journal]{IEEEtran}
\usepackage{times}
\usepackage{epsfig}
\usepackage{graphicx}
\usepackage{cite}
\usepackage{amsmath,amssymb,amsfonts}
\usepackage{algorithmic}
\usepackage{graphicx}
\usepackage{textcomp}
\usepackage{xcolor}
\usepackage{threeparttable}
\usepackage{subfigure}
\usepackage{booktabs}
\usepackage[misc]{ifsym}
\usepackage{algorithm}  
\usepackage{algorithmic} 
\usepackage{footnote}
\usepackage{color}
\usepackage{eqparbox,array}
\usepackage{multirow}
\usepackage{bbding} %

\usepackage[pagebackref=true, colorlinks,
            linkcolor=red,
            anchorcolor=black,
            citecolor=black
            ]{hyperref}

\def\cvprPaperID{****} % *** Enter the CVPR Paper ID here

\begin{document}

%%%%%%%%% TITLE
\title{Prototypical Contrast and Reverse Prediction: Unsupervised   Skeleton Based Action Recognition}

\author{Shihao Xu, Haocong Rao, 
 Xiping Hu,  Bin Hu
        % <-this % stops a space
\thanks{
% This work was supported in part by ...
\textit{Corresponding authors: Xiping Hu;  Bin Hu.}}% <-this % stops a space

}

% \and

\maketitle
%\thispagestyle{empty}

%%%%%%%%% ABSTRACT
\begin{abstract}
In this paper, we focus on unsupervised representation learning for skeleton-based action recognition.
Existing approaches usually learn action representations by sequential prediction but they suffer from the inability to fully learn semantic information. To address this  limitation, we propose a novel framework   named \textbf{P}rototypical \textbf{C}ontrast and \textbf{R}everse \textbf{P}rediction (PCRP),
which not only creates reverse sequential prediction to learn low-level information (\textit{e.g.,} body posture at every frame) and high-level pattern (\textit{e.g.,} motion order), 
but also devises action prototypes to    implicitly encode   semantic similarity shared among  sequences.  
In general, we regard action prototypes  as latent variables and formulate PCRP as an expectation-maximization  task. 
Specifically, PCRP iteratively runs (1) E-step as determining the distribution of prototypes by clustering  action encoding from the encoder, and (2) M-step as optimizing the encoder by minimizing the proposed ProtoMAE loss, 
which helps simultaneously pull  the action encoding closer to its assigned prototype and perform reverse prediction task.  Extensive experiments on N-UCLA, NTU 60, and NTU 120 dataset present that PCRP outperforms  state-of-the-art unsupervised methods and   even achieves superior performance over some of supervised methods. 
Codes are available at \href{https://github.com/Mikexu007/PCRP}{https://github.com/Mikexu007/PCRP}.
\end{abstract} 

%%%%%%%%% BODY TEXT
\section{Introduction}
As an essential branch in computer vision, skeleton based action recognition has drawn broad  attention due to the  compact  and effective skeletal representation of human body and its robustness against viewpoint variations  and noisy backgrounds \cite{si20enhanced19an, Liang_2019_CVPR_Workshops, shi2019skeletonbased, Cheng_2020_CVPR}.

Many of current skeleton-based works \cite{zhang2017view,yan2018spatial,Cheng_2020_CVPR} for action recognition  resort to supervised learning paradigms to  learn action representations, which require  massive annotated  samples for training. However, the  annotated information  sometimes is not available or  demand expensive labor force for labelling, which might face uncertain labelling or mislabelling challenges due to the high inter-class  similarity of actions \cite{devillers2005challenges,wang2006annosearch}. From this perspective, exploiting the unlabeled  data to learn effective action  representations  arouses considerable interests \cite{kulkarni2019unsupervised,nguyen2019hologan}. 

\begin{figure}[t]
    \centering
    % 0.52
    \scalebox{0.45}{
    \includegraphics{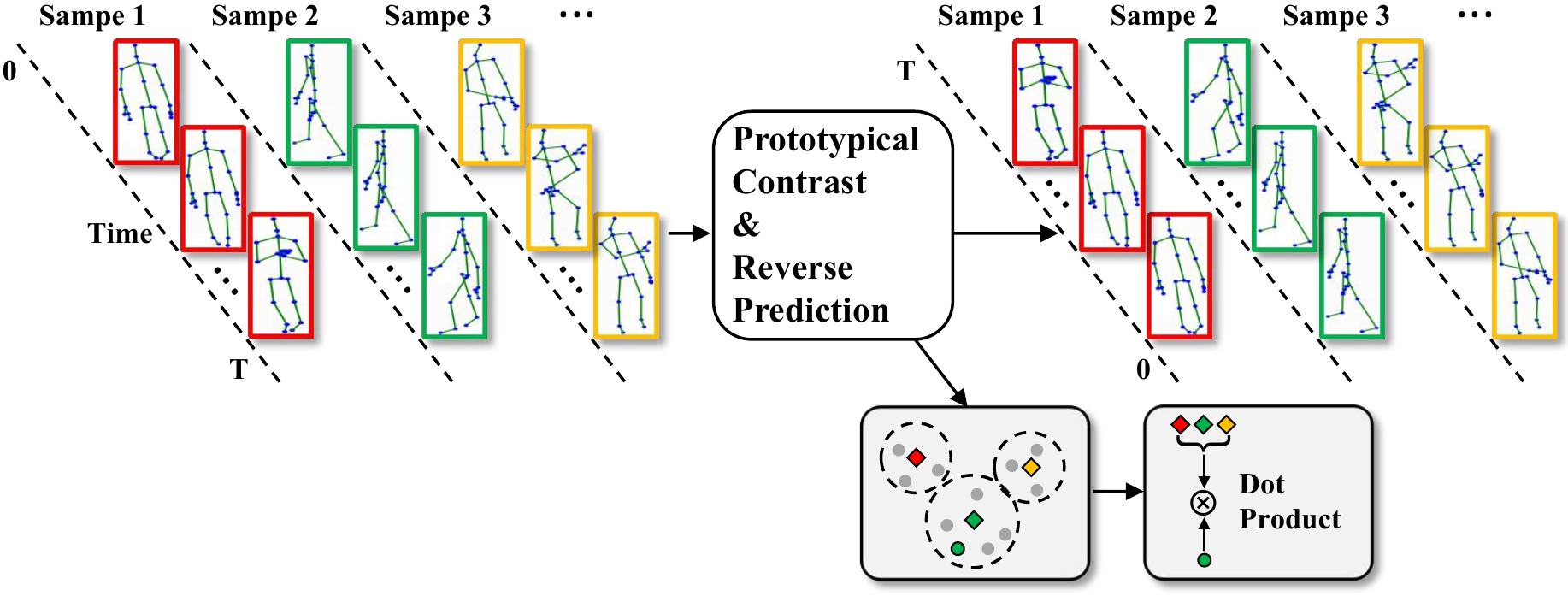}    }
    \caption{Illustration of Prototypical Contrast and Reverse Prediction framework.
     } 
    \label{fig:first}
\end{figure}

In recent years, a stream of unsupervised   learning  methods  have been introduced. 
%%%%%% here add new studies acm mm
Most  of them \cite{luo2017unsupervised, zheng2018unsupervised,su2020predict,lin2020ms2l, rao2020augmented} are built  upon  encoder-decoder structures \cite{badrinarayanan2017segnet} to yield discriminative action representations  via sequential prediction/reconstruction  or augmented sequence contrast.
% while the other work \cite{rao2020augmented}  not only obtain effective representations but also achieve some invariance against skeleton  transformation through instance contrasting tasks \cite{wu2018unsupervised,He_2020_CVPR,chen2020simple}.  
However, these methods suffer from  a common significant disadvantage:  Higher-level semantics (\textit{e.g.,} motion order, semantic similarity among sequences) is not fully explored. This issue derives from the instance-level  situation that the sequential prediction task  forces the predicted sequence to  get closer to only the original  one, but neglect the semantic similarity between various instances. Likewise, augmented sequence contrast is also restricted in  pulling closer  two augmented samples of one sequence regardless of others.   
Furthermore,  this problem is worsened in large-scale datasets,  since  the  correlation  shared among numerous semantically similar samples cannot be fully  exploited.

% Furthermore, when  involved in large-scale datasets, this problem is worsened by the fact that numerous samples sharing similar semantics  are  used for  predicting only themselves respectively or for pair-wise matching of their augmented instances derived from an individual,  leading to   considerable but undesirable  loss of semantic information. 
% On the other hand,  predicting data in order seems like a stereotype to follow, and yet sequence prediction in reverse order is rarely explored.
% and the instance contrasting task tries to pull closer samples from the same instance but push apart samples from different instances. 
% Apparently they both 

To address the challenges above,   we rethink the encoder-decoder based sequential prediction in terms of expetation-maximization (EM) algorithm \cite{dempster1977maximum}, and propose \textbf{P}rototypical \textbf{C}ontrast and \textbf{R}everse \textbf{P}rediction (PCRP) framework.
Fig.\ref{fig:first} illustrates the proposed PCRP. 
An action prototype, similar to an image prototype \cite{li2020prototypical}, is a representative encoding for a bunch of semantically similar sequences.
Instead of directly using encoder-decoder structure to obtain  representation via data prediction,
we exploit the EM algorithm to  encode semantic structure of data into action representations by 
(1) implicitly learning  semantic similarity between  sequences  to force the action encoding to approach their corresponding prototypes, 
and (2) learning high-level information (\textit{e.g.,} motion order) of sequences via predicting sequence in reverse order. 

\begin{figure*}[t]
    \centering
    % 0.52
         \subfigure[Illustration of PCRP in view of EM algorithm. In E-step, the action encoding from the final output of the encoder is used for  clustering. In M-step, the action encoding is fed into the decoder for predicting sequence reversely, and   ProtoMAE loss is minimized   to update the  encoder. BP denotes back propagation.]{\scalebox{0.39}{
    \includegraphics{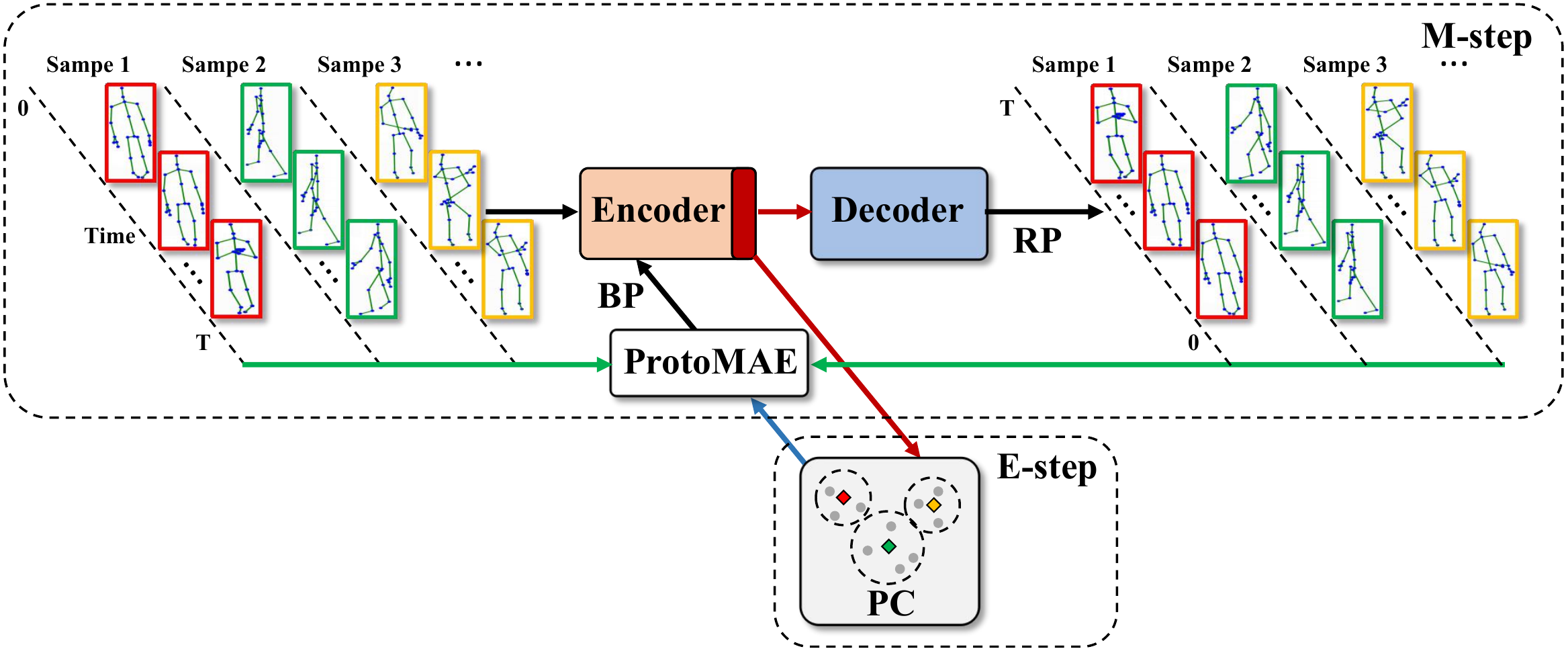}}
    \label{fig:framework}}
        \quad
         \subfigure[An action encoding can be assigned to different action prototypes with different granularity. PCRP attempts to pull the encoding closer to the most suitable prototype. ]{\scalebox{0.39}{
    \includegraphics{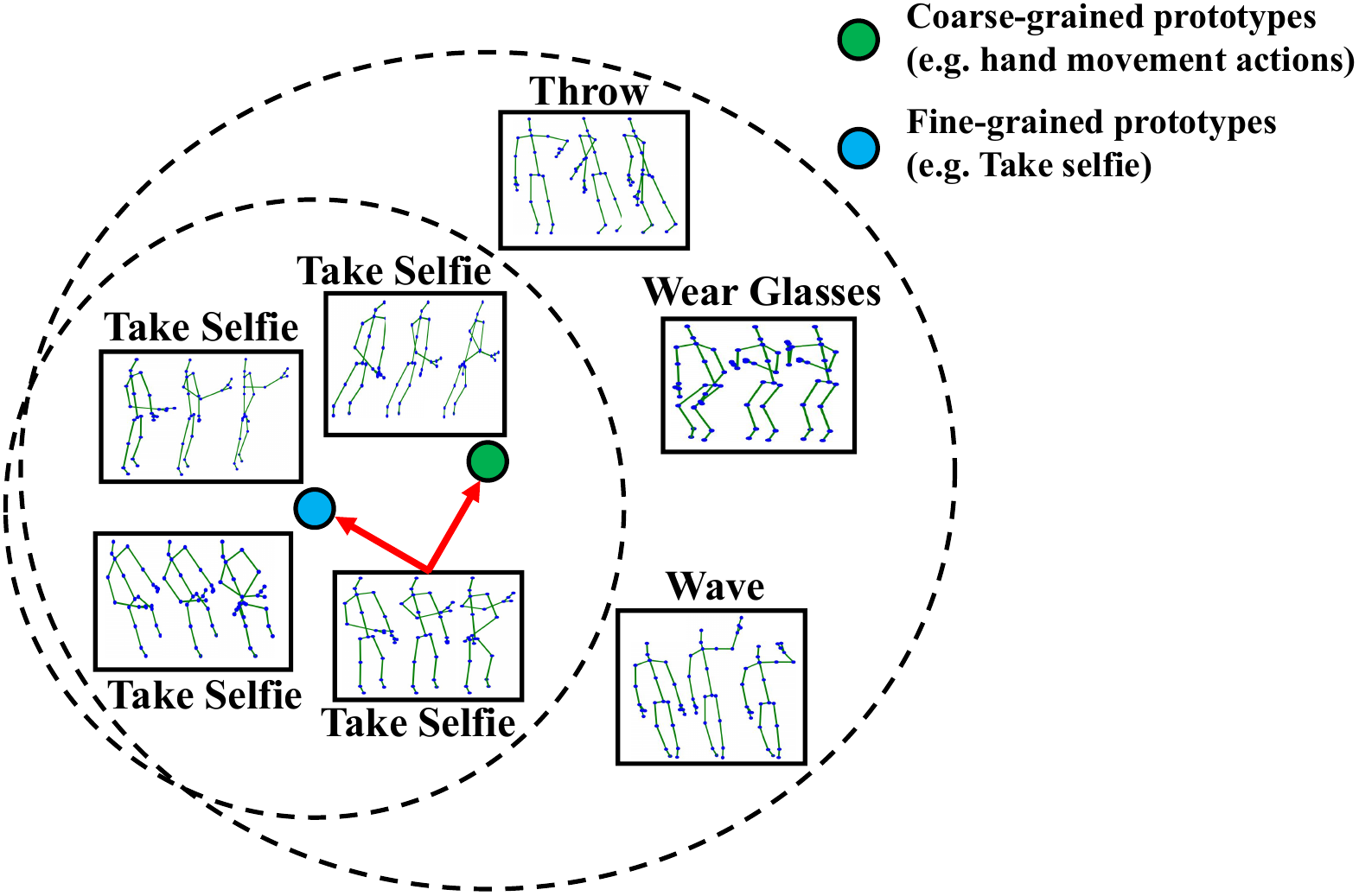}}
    \label{fig:two_circles}}
    \caption{}
\quad
\end{figure*}

Specifically, we focus on  the encoder parameter learning in the EM algorithm and regard action prototypes as additional latent variables.
From this perspective,  the EM algorithm attempts to find a maximum likelihood estimate of encoder parameters (see Fig. \ref{fig:framework}),  while the decoder  keeps fixed   for enhancing the encoder to learn representations \cite{su2020predict}. Given the current encoder parameters, the expectation step (E-step) aims to estimate the probability of prototypes by performing \textit{k}-means clustering on the action encoding (the output at final step) from the Uni-GRU encoder, and the maximization step (M-step) tries to update the encoder parameters by minimizing the proposed loss, namely, ProtoMAE (Sec. \ref{subsubsection:protomae}). 
Minimizing ProtoMAE is  equivalent to   maximizing  the estimated likelihood under the assumption that the distribution around each protoype is isotropic Gaussian \cite{li2020prototypical}.  
It  is also equivalent to help   predict  sequence reversely and simultaneously pull the action encoding closer to its corresponding prototype compared to other prototypes (see Fig. \ref{fig:two_circles}). 
The E-step and the M-step function iteratively. 
In this way, the encoder is able to learn discriminative action representaions without labeled data, and after convergence, it can be used for other downstream tasks such as  classification. 
The contributions of our work are listed as follows:

\begin{itemize}
    \item We propose a novel framework named Prototypical Contrast and Reverse Prediction to explore high-level information of sequences and  that of the global dataset. To our knowledge, this work is the first to introduce prototypical contrast and reverse prediction  for unsupervised skeleton based action recognition.
    
    \item We formulate the PCRP into an EM iteration manner, in which the alternating steps of clustering and reverse  prediction  serve to approximate and maximize the log-likelihood function.
    
    \item 
    We  introduce  ProtoMAE, an enhanced MAE loss that exploits contrastive loss to achieve high-level information learning as well as to  adaptively estimate the tightness of the feature distribution around each prototype. 
    
    \item Experiments on the N-UCLA, NTU RGB+D 60, and NTU RGB+D 120 dataset, show the superiority of our framework to other state-of-the-art unsupervised methods as well as some of supervised counterparts.
\end{itemize}

\section{Related Work}
% \textbf{Skeleton based Action Recognition:}
% Most of \textit{supervised} deep neural networks (DNN) based models  have achieved great success in skeleton based action recognition. 
% For example, 
% Long Short Term (LSTM) networks, that are powerful to model temporal action dynamics, are usually adopted and modified to extract informative joint features \cite{shahroudy2016ntu,liu2017skeleton1} or establish  attention enhancements \cite{liu2017skeleton, si20enhanced19an}. Convolutional neural network (CNN) based models \cite{ke2017new, liu2017enhanced, ke2018learning, liu2018recognizing, Liang_2019_CVPR_Workshops} are often used to extract spatial and temporal motion features when skeletons are transformed into pseudo-images. 
% Recently, by exploiting the attributes of human topology, numerous graph convolutional network (GCN) based models are able  to  capture more distinguished spatio-temporal features with directed graph \cite{Shi_2019_CVPR}, adaptive mechanism \cite{shi2019two, shi2019skeletonbased} or lightweight design \cite{Cheng_2020_CVPR}. 

\textbf{Unsupervised action Recognition:}
While supervised methods \cite{Liang_2019_CVPR_Workshops, Shi_2019_CVPR, Cheng_2020_CVPR} show great performance in skeleton based action recognition by using annotated information, \textit{unsupervised}  methods are advantageous at learning action  representation without any labels.  
Zheng \textit{et al.} \cite{zheng2018unsupervised} introduce a generative adversarial network (GAN) based encoder-decoder for skeletal sequence regeneration, and utilize the representation learned from encoders to identify actions. 
Su \textit{et al.} \cite{su2020predict} further devise predict\&cluster (P\&C) model with decoder-weakening mechanism to enhance the  ability of the encoder to capture more discriminative action pattern. 
Rao \textit{et al.} \cite{rao2020augmented} propose skeleton augmentation strategies and apply momentum LSTM with contrastive learning to learn robust action representation. However, these methods ignore the semantic information between different sequences.
In this paper, we adopt encoder-decoder structure with decoder-weakening strategy \cite{su2020predict} as the backbone,  and  propose prototypical  contrast for  semantic learning and achieve sequential reverse prediction for enhancing representation learning.

\textbf{Unsupervised Action Clustering:} 
Many clustering based models have been introduced for unsupervised action clustering. Jones \textit{et al.} \cite{Jones_2014_CVPR} propose dual assignment k-means (DAKM) to achieve context learning for facilitating unsupervised  action clustering. Bhatnagar \textit{et al.} \cite{bhatnagar2017unsupervised} devise weak learner based autoencoders to extract temporal features under different temporal resolutions. Peng  \textit{et al.} \cite{peng2019recursive} establish a recursive constrained model by  using the contextual motion and scene for unsupervised video action clustering. Nevertheless, these approaches only serve for RGB videos and yet the counterpart for skeleton action sequences is not  developed. In this proposed work, we for the first time explore the prototypical contrast for unsupervised skeleton based action recognition.

\textbf{Contrastive Learning:}
In recent years, contrastive learning, a type of unsupervised (self-supervised) learning method, has attracted massive attention. Most of them  \cite{ He_2020_CVPR, chen2020simple, caron2020unsupervised, li2020prototypical} learn effective representations by  pretext tasks \cite{wu2018unsupervised,zhuang2019local} with contrastive losses \cite{hadsell2006dimensionality,gutmann2010noise}.  For example, Wu \textit{et al.} \cite{wu2018unsupervised} base an instance contrast task and noise-contrastive  estimation  (NCE)  loss  \cite{gutmann2010noise}  to match positive pairs and push apart negative pairs. He \textit{et al.} \cite{He_2020_CVPR} propose momentum based encoder to learn more consistent  representations. Nevertheless,  these methods mainly focus on image representation learning. In this paper, we introduce prototypical contrast \cite{li2020prototypical} to skeleton based action recognition and improve the sequential prediction task on high-level semantics learning.

% \section{Method}
\section{Preliminaries}
We focus on the unsupervised representation learning using skeleton sequences. Then, we exploit the learned representations for skeleton-based action recognition. 
Given a training set $\Phi = \left\{\boldsymbol{x}^{(i)}\right\}_{i=1}^{N}$ of $N$ skeleton sequences,  each  sequence $\boldsymbol{x} \in \mathbb{R}^{T \times J \times 3}$ contains $T$ skeleton frames and each frame has $J$ body joints that are represented in 3D space. 
Our goal is to learn an  encoder $f_{E}$ (we employ Uni-GRU) that maps $\Phi$ to action encoding set $V=\left\{\boldsymbol{v}^{(i)}\right\}_{i=1}^{N}$, where  $\boldsymbol{v}^{(i)} \in \mathbb{R}^{C}$   is a discriminative action representation of  $\boldsymbol{x}^{(i)}$. 
Traditional  encoder-decoder based models achieve this goal by sequential prediction as to optimize the loss function of mean square error (MSE) or mean absolute error (MAE) between the original   sequence and its predicted one. 
% But they do not consider semantic learning.  We take MAE as an example and represent it as follows:
% \begin{equation}
% \mathcal{L}=\frac{1}{T}\frac{1}{J} \sum_{t=1}^{T}\sum_{j=1}^{J}\left|\boldsymbol{x}_{t,j}-\hat{\boldsymbol{x} }_{t,j}\right|
% \end{equation}
% where $\hat{\boldsymbol{x} }_{t,j}$ is predicted position of joint $j$  at $t^{th}$ frame from decoders.
MAE/MSE only focus on skeleton reconstruction within each single sequence and ignore the similarity of different sequences. In our proposed framework PCRP, we tackle this challenge by introducing action prototypical contrast paradigm (see Sec. \ref{sec:proto clustering}). Besides, we  achieve sequential  prediction in reverse order (see Sec. \ref{sec:reverse prediction}) to enhance high-level information (\textit{e.g.,} motion pattern) learning. Fig. \ref{fig:framework} illustrates our framework, where semantic learning and data reverse prediction are performed alternately at each epoch. 
% Next, we elaborate our PCRP framework from the angle of  EM. 
The  main algorithm of PCRP is shown in Algorithm \ref{algorithm1}.

Before introducing our proposed PCRP, we first have a brief review of the general encoder-decoder based sequential prediction task that we rely on.

\subsection{Sequential Prediction}
\label{sec:seq pred}
Given a skeleton sequence $\boldsymbol{x} = \left\{\boldsymbol{x}_1, \ldots, \boldsymbol{x}_T \right\}$, the model is expected to output the predicted sequence  $\hat{\boldsymbol{x}}=\left(\hat{\boldsymbol{x}}_1, \ldots, \hat{\boldsymbol{x}}_T \right)$ that gets closer as much as possible to  $\boldsymbol{x}$. 
In training phase, the encoder (\textit{e.g.,} Uni-GRU) encodes every skeleton frame $\boldsymbol{x}_t$ ($t \in \left \{1, \ldots T \right\}$) and the previous step's latent state $\boldsymbol{h}_{t-1}$ ($t - 1 > 0$) to determine the current output $\boldsymbol{v}_t$  and the current latent state $\boldsymbol{h}_t$:
\begin{equation}
\left(\boldsymbol{v}_t, \boldsymbol{h}_{t}\right)
=\left\{\begin{array}{ll}
f_{E}\left(\boldsymbol{x}_t\right) & \text { if } t=1 \\
f_{E}\left(\boldsymbol{h}_{t-1}, \boldsymbol{x}_t\right) & \text { if }  t>1
\end{array}\right.
\end{equation}
where $\boldsymbol{v}_t, \boldsymbol{h}_{t} \in \mathbb{R}^{C}$. Next, the decoder $f_D$ utilizes  the  output at final step $\boldsymbol{v}_T$  from the encoder to perform prediction task:
\begin{equation}
\left(\hat{\boldsymbol{x}}_{t}, \hat{\boldsymbol{h} }_{t}\right)=\left\{\begin{array}{ll}
f_{D}\left(\boldsymbol{v}_{T} \right) & \text { if }  t=1 \\
f_{D}\left(\hat{\boldsymbol{h}}_{t-1}\right) & \text { if }  t>1
\end{array}\right.
\end{equation}
Then MAE loss is  applied on $\boldsymbol{x}$ and   $\hat{\boldsymbol{x}}$ for model optimization.
Therefore, $\boldsymbol{v}_T$ is the action encoding (\textit{i.e.,} representation) of the sequence $\boldsymbol{x}$.

 \begin{algorithm}[!t]
	\caption{Main algorithm of PCRP}%算法标题
	 \label{algorithm1}
	\begin{algorithmic}%一行一个标行号
	\footnotesize 
    \STATE \textbf{Input:} encoder $f_E$, decoder $f_D$, training dataset $\Phi$, number of clusters $K = \left\{k_m\right\}^M_{m=1}$

    \WHILE{not MaxEpoch} 
    \STATE \textcolor[rgb]{0,0.4,1}{\# E-step}
    
     \STATE $V=f_E(\Phi)$
       \textcolor{gray}{\COMMENT{obtain  action encoding for all training data} }
       
    \FOR{$m = 1$ \textbf{to} $M$}
    \STATE \textcolor{gray}{\# cluster $V$ into $k_m$ clusters and return prototypes.}
    \STATE $\boldsymbol{Z}^{m}=k\text{-means}\left(V, k_{m}\right)$ 

    \STATE \textcolor{gray}{\# calculate the distribution tightness of each prototype with Eq. \ref{eq:tightness}}
    
    \STATE $\phi_{m}=$ Tightness $\left(\boldsymbol{Z}^{m}, V\right)$
    
    \ENDFOR
     
     \STATE \textcolor[rgb]{0,0.4,1}{\# M-step}
     \FOR{a  mini-batch $\boldsymbol{x}$ \textbf{in} $\Phi$}
     \STATE $\boldsymbol{v}=f_E(\boldsymbol{x})$ 
     
     \STATE $\boldsymbol{\hat{x}}=f_D(\boldsymbol{v})$  
     
     \STATE $\overline{\boldsymbol{x}}=\text{Reverse}(\boldsymbol{x})$
     
     \STATE \textcolor{gray}{\# compute loss with Eq.\ref{eq:protomae}}
     
     \STATE $\mathcal{L}_{\text{ProtoMAE}} \left(\boldsymbol{v}, \overline{\boldsymbol{x}}, \boldsymbol{\hat{x}},\left\{\boldsymbol{Z}^{m}\right\}_{m=1}^{M},\left\{\phi_{m}\right\}_{m=1}^{M}\right)$

     \STATE fix $f_D$ \textcolor{gray}{\COMMENT{parameters of decoder do not evolve} }
     
     \STATE Update $f_E$ to minimize $\mathcal{L}_{\text{ProtoMAE}}$ with Adam optimizer
     
     \ENDFOR
     
    \ENDWHILE
	\end{algorithmic}
\end{algorithm}

\section{Prototypical Contrast and Reverse Prediction as Expectation-Maximization}

Sequence prediction based PCRP  aims to find the encoder parameters $\boldsymbol{\theta}$ that maximizes the likelihood function of the  $N$ observed sequences: 
\begin{equation}\boldsymbol{\theta}^{*}=\underset{\boldsymbol{\theta}}{\arg \max } \sum_{i=1}^{N} \log p\left(\boldsymbol{x}^{\left(i\right)} \mid
\boldsymbol{\theta}\right)
\label{eq:argmax}
\end{equation}.

Since the action prototypes are  introduced but not directly observed,  they are  viewed as the latent variables  of observed data given by $\boldsymbol{Z} = \left\{\boldsymbol{z}_i\right\}_{i=1}^{K}$ with $K$ action prototypes, where $\boldsymbol{z}_i \in \mathbb{R}^C$. Thus the Eq. \ref{eq:argmax} is referred to as:
\begin{equation}\boldsymbol{\theta}^{*}=\underset{\boldsymbol{\theta}}{\arg \max } \sum_{i=1}^{N} \log \sum_{\boldsymbol{z}_{i} \in \boldsymbol{Z} } p\left(\boldsymbol{x}^{\left(i\right)}, \boldsymbol{z}_{i} \mid \boldsymbol{\theta}\right)  \label{eq:argmax2}.\end{equation}
Achieving this function directly is challenging, and the only knowledge of action prototypes $\boldsymbol{Z}$ is obtained in the posterior distribution  $ p(\boldsymbol{z}_i\mid\boldsymbol{x}^{\left(i\right)},\boldsymbol{\theta})$.
Under this circumstance,
we first utilizes  current parameters $\boldsymbol{\theta}^{\text{old}}$ and   the Jensen's inequality  to turn Eq. \ref{eq:argmax2} into an  expectation\footnote{More details are given in Supplementary Materials.}  
$\mathcal{Q}\left(\boldsymbol{\theta}, \boldsymbol{\theta}^{\mathrm{old}}\right)$
% of 
% $ \log p(\boldsymbol{x}^{\left(i\right)}, \boldsymbol{z}_i \mid \boldsymbol{\theta})$  
that needs  to be maximized:
\begin{equation}
\boldsymbol{\theta}^{*}=\underset{\boldsymbol{\theta}}{\arg \max } \mathcal{Q}\left(\boldsymbol{\theta}, \boldsymbol{\theta}^{\mathrm{old}}\right)
\label{eq:argmax3},
\end{equation}
% ===========
\begin{equation}
\begin{small}
\mathcal{Q}\left(\boldsymbol{\theta}, \boldsymbol{\theta}^{\mathrm{old}}\right)=\sum_{i=1}^{N} \sum_{\boldsymbol{z}_{i}\in\boldsymbol{Z}} p\left(\boldsymbol{z}_i\mid\boldsymbol{x}^{\left(i\right)}, \boldsymbol{\theta}^{\mathrm{old}}\right) \log p(\boldsymbol{x}^{\left(i\right)}, \boldsymbol{z}_i \mid \boldsymbol{\theta}).
\end{small}
\label{eq:expectation}
\end{equation}
Then we rely on  the EM algorithm with  E-step and M-step  to achieve Eq. \ref{eq:argmax3}. 
% ========== ========== E step
\subsection{E-step}
In this step,   we attempt to estimate 
$p\left(\boldsymbol{z}_{i} \mid \boldsymbol{x}^{(i)}, \boldsymbol{\theta}^{\text {old }}\right)$ of Eq. \ref{eq:expectation} and introduce prototypical contrast.

% ==================== Prototypical contrast
\subsubsection{Prototypical Contrast}
\label{sec:proto clustering}
The result of $p\left(\boldsymbol{z}_{i} \mid \boldsymbol{x}^{(i)}, \boldsymbol{\theta}^{\text {old }}\right)$ 
is based on the action prototype $\boldsymbol{z}_i$. 
Along this line, we take advantage of the action encoding  from encoder  to obtain  $\boldsymbol{z}_i$. 
Specifically, we  apply \textit{k}-means  algorithm on all  action encoding $\left\{\boldsymbol{v}^{(i)}_T\right\}_{i=1}^{N}$   (the  final output) from $f_E$ to obtain $K$ clusters, in which we  define prototype $\boldsymbol{z}_i \in \mathbb{R}^C$ as the centroid of the $i^{th}$ cluster \cite{li2020prototypical}. 
Therefore, we have
\begin{equation}
p\left(\boldsymbol{z}_{i} \mid \boldsymbol{x}^{(i)}, \boldsymbol{\theta}^{\text {old }}\right) = 
\left\{\begin{array}{ll}
0 & \text { if } \boldsymbol{v}^{(i)}_T \notin \boldsymbol{z}_i  \\
1 & \text { if }  \boldsymbol{v}^{(i)}_T \in \boldsymbol{z}_i
\end{array}\right..
\label{eq:0 or 1}
\end{equation}
Using the action encoding from encoder to achieve prototypical contrast is beneficial due to several aspects: (1) The action encoding is in low dimension compared with the whole sequence. (2) The action encoding contains abundant context information of the action. (3) Semantic similarity   between different samples is explored  by   pulling the action encoding closer to their corresponding prototypes (see Sec. \ref{subsubsection:protomae}).  

% ============================ Tightness estimation
\subsubsection{Tightness Estimation}
To measure the cluster's quality (feature distribution), we introduce the tightness   $\phi \propto \sigma\footnote{$\sigma$ denotes standard deviation of data distribution}$ \cite{li2020prototypical}.
% in which we expect smaller $\phi$ shows  larger tightness of each cluster. 
We first suppose a cluster has a  prototype $\boldsymbol{z}_i$ and contains $P$ action encoding vectors  $\left\{\boldsymbol{v}^{(i)}_T\right\}_{i=1}^{P}$, 
which are then used to compute $\phi$. Here  a good  $\phi$ is expected to be small and  satisfy several requirements:  (1)  The average distance between each action encoding $\boldsymbol{v}^{(i)}_T$ and their prototype $\boldsymbol{z}_i$ is  small. (2) A  cluster covers more action encoding (\textit{i.e.,} $P$ is large). 
To achieve this goal, we define  $\phi$ as follows:
\begin{equation}
\phi=\frac{\sum_{i=1}^{P}\left\|\boldsymbol{v}^{(i)}_T-\boldsymbol{z}_i\right\|_{2}}{P \log (P+\alpha)},
\label{eq:tightness}
\end{equation}
where $\alpha$ is a scaling parameter that avoids overwhelmingly large $\phi$.  On the other hand, $\phi$ serves as a punishing factor in the loss objective (see Sec. \ref{sec:phi explain}) to generate more balanced clusters with similar tightness.

%  f_{E}\left(\boldsymbol{x}^{(i)}\right) 
% utilize the current parameter $\boldsymbol{\theta}^{\text{old}}$ to determine the posterior distribution of $\boldsymbol{Z}$ given by $ p(\boldsymbol{z}_i\mid\boldsymbol{x}^{\left(i\right)},\boldsymbol{\theta})$,
% with which we then find the expectation of the surrogate likelihood $\mathcal{Q}\left(\boldsymbol{\theta}, \boldsymbol{\theta}^{\text {old }}\right)$ represented as: 
% ========== M step
\subsection{M-step}
%, the renewed parameter $\boldsymbol{\theta}^{\text{new}}$ is generated by maximizing the below function:
% \begin{equation}\boldsymbol{\theta}^{\text {new }}=\underset{\boldsymbol{\theta}}{\arg \max } \mathcal{Q}\left(\boldsymbol{\theta}, \boldsymbol{\theta}^{\text {old }}\right).\end{equation}
Next, we try to estimate 
$ p\left(\boldsymbol{x}^{(i)}, \boldsymbol{z}_{i} \mid \theta\right)$. 
Due to the uniform probability over cluster centroids, we set  $p\left(\boldsymbol{z}_{i} \mid \boldsymbol{\theta}\right) = \frac{1}{K} $ and get:
\begin{align}
p\left(\boldsymbol{x}^{(i)}, \boldsymbol{z}_{i} \mid \boldsymbol{\theta}\right) &=p\left(\boldsymbol{x}^{(i)} \mid  \boldsymbol{z}_{i}, \boldsymbol{\theta}\right) p\left(\boldsymbol{z}_{i} \mid  \boldsymbol{\theta}\right) \notag \\ &=  \frac{1}{K} \cdot p\left(\boldsymbol{x}^{(i)} \mid  \boldsymbol{z}_{i}, \boldsymbol{\theta}\right). \label{eq:1/k}
\end{align}
To calculate Eq. \ref{eq:1/k}, we  assume that the distribution for each action prototype is an isotropic Gaussian \cite{li2020prototypical}, which results in:
\begin{equation}
\begin{small}
p\left(\boldsymbol{x}^{(i)} \mid \boldsymbol{z}_{i}, \boldsymbol{\theta}\right)=
\frac{\exp \left(\frac{-\left(\boldsymbol{v}^{(i)}_T-\boldsymbol{z}_{s}\right)^{2}}{2 \sigma_{s}^{2}}\right) }{\sum_{k=1}^{K} \exp \left(\frac{-\left(\boldsymbol{v}^{(i)}_T-\boldsymbol{z}_{k}\right)^{2}}{2 \sigma_{k}^{2}}\right)}, \label{eq: iso gau}
\end{small}
\end{equation}
where $\boldsymbol{v}^{(i)}_T \in \boldsymbol{z}_{s}$. Suppose $\ell_{2}$-normalization is applied to $\boldsymbol{v}^{(i)}_T$ and $\boldsymbol{z}_i$, then we  have $(\boldsymbol{v}^{(i)}_T-\boldsymbol{z}_i)^{2}=2-2 \boldsymbol{v}^{(i)}_T \cdot \boldsymbol{z}_i$.
On the basis of Eq. \ref{eq:argmax3}, \ref{eq:expectation}, \ref{eq:0 or 1}, \ref{eq:1/k}, \ref{eq: iso gau}, the maximum likelihodd estimation is referred to as:
\begin{equation}
\boldsymbol{\theta}^{*}=
\underset{\boldsymbol{\theta}}{\arg \min } \sum_{i=1}^{N}-\log \frac{\exp \left(\boldsymbol{v}^{(i)}_T \cdot \boldsymbol{z}_{s} / \phi_{s}\right)}{\sum_{k=1}^{K} \exp \left(\boldsymbol{v}^{(i)}_T \cdot \boldsymbol{z}_{k} / \phi_{k}\right)},
\label{eq:new likelihood}
\end{equation}
Note that Eq. \ref{eq:new likelihood} is  a kind of contrastive loss (similar as InfoNCE \cite{oord2018representation}), which evaluates the affinity between  the action encoding and its assigned prototype over the affinity between that action encoding and other prototypes. 

Based on  Eq. \ref{eq:new likelihood}, we further introduce sequential reverse prediction and add the related MAE loss to help preserve low-level information that can regenerate the sequence.  Thus we  construct the overall objective, namely ProtoMAE (see Sec. \ref{subsubsection:protomae}).

% ============ RC
\subsubsection{Reverse Prediction}
\label{sec:reverse prediction}
Instead of performing commonly-used plain sequential prediction (see Sec. \ref{sec:seq pred}) for action representation learning, we propose reverse prediction as to learn more high-level information (\textit{e.g.} movement order) that are meaningful to human perception. 
Hence, we expect our model is able to generate predicted sequence $\hat{\boldsymbol{x}}=\left(\hat{\boldsymbol{x}}_1, \ldots, \hat{\boldsymbol{x}}_T \right) $ that get closer to  
$\overline{\boldsymbol{x}}=\left\{\overline{\boldsymbol{x}}_1, \ldots, \overline{\boldsymbol{x}}_T \right\}=\left\{\boldsymbol{x}_T, \ldots, \boldsymbol{x}_1 \right\}$, 
where 
$\overline{\boldsymbol{x}}_t = \boldsymbol{x}_{T-t+1}$. Then the MAE loss for reverse prediction is  defined as: 
\begin{equation}
    \mathcal{L}_R=\frac{1}{T}\frac{1}{J} \sum_{t=1}^{T}\sum_{j=1}^{J}\left|\overline{\boldsymbol{x}}_{t,j}-\hat{\boldsymbol{x} }_{t,j}\right|.
    \label{eq:new mae}
\end{equation}

\subsubsection{ProtoMAE Loss}
\label{subsubsection:protomae}
To this end,  we  combine Eq.\ref{eq:new mae} and  Eq. \ref{eq:new likelihood} to form a new loss objective named ProtoMAE, defined as:

\begin{align}
\begin{scriptsize}
\mathcal{L}_{\text {ProtoMAE }} =\sum_{i=1}^{N} \left( \sum_{t=1}^{T}\left|\overline{\boldsymbol{x}}_{t}-\hat{\boldsymbol{x} }_{t}\right| 
% \notag \right.
% \\
% \phantom{=\;\;}
% \left.
- \frac{1}{M} \sum_{m=1}^{M} \log \frac{\exp \left(\frac{\boldsymbol{v}^{(i)}_T \cdot \boldsymbol{z}_{s}^m }{\phi_{s}^m} \right)}{\sum_{k=1}^{r} \exp \left( \frac{\boldsymbol{v}^{(i)}_T \cdot \boldsymbol{z}_{k}^m }{\phi_{k}^m} \right)}   \right), 
\end{scriptsize}
\label{eq:protomae}
\end{align}
which is to be minimized to simultaneously achieve sequential reverse prediction and cluster the action encoding with semantic similarity.
\label{sec:phi explain}
Note that in Eq. \ref{eq:protomae} large  $\phi$ denotes the action encoding are in a loose cluster and small  $\phi$ means they are in a tight cluster.  Large  $\phi$ weakens the affinity between the action encoding and the prototype, which  drives the encoder to pull the action encoding closer to the prototype. In contrast, small $\phi$ does not compromise much to the affinity mentioned above, which less encourages  the action encoding approach  the prototype. 
Hence, learning with ProtoMAE generates more balanced clusters with similar tightness \cite{li2020prototypical}.
Besides, since the $K$ may be too large, we choose to sample $r$  prototypes, where $r$ \textless $K$.  We also attempt to cluster action encoding $M$ times  with different number of clusters 
$K=\left\{k_{m}\right\}_{m=1}^{M}$ to provide more robust probability estimation of prototypes.

EM algorithm performs E-step and M-step   alternately without supervision for a specific epochs. 
Then the  quality of learned representations $\boldsymbol{v}_T$ from the encoder are measured by linear evaluation protocol\cite{zheng2018unsupervised}, where the learned representations are always kept frozen and a linear classifier is added on top of  them for training and testing.

% \section{Practice of PCRP}
% In practice, the PCRP consists of three operations, including prototype estimation ($O_{P}$), tightness estimation ($O_{C}$), and likelihood maximization ($O_M$). Briefly, 
% % given the training set $\Phi $ and each sequence $\boldsymbol{x} \in \mathbb{R}^{T \times D}$, where $D = J \times 3$,   
% $O_{P}$ bases on the training set and the current encoder parameters to estimate the latent variables, namely, action prototypes $\boldsymbol{Z} \in \mathbb{R}^{K \times C}$ where $K$ is the number of prototypes and $C$ is the feature dimension. $O_{C}$ further estimates the tightness $\phi$  of action encodings around every prototype as a  scaling factor in  Eq. \ref{eq:new likelihood}. 
% Thus $O_{P}$ and $O_{C}$ work as E-step in the EM algorithm. $O_{M}$  combines Eq. \ref{eq:new likelihood} and the MAE loss to form a new loss function ProtoMAE (See Sec. \ref{}), which is to be minimized  to  achieve reverse prediction and prototypical clustering simultaneously. So $O_{M}$ is the M-step in EM.  To this end,  $O_{P}$, $O_{C}$, $O_{M}$ run iteratively to learn  action representation without supervision for a specific epochs. 

\section{Experiments}
\textbf{Dataset:}  Experiments are based on three large action datasets 
% and they are all captured by Kinect cameras, which generate RGB, depth, and skeletons information of human actions. In this work, 
and we use their skeleton sequences.
(1) \textbf{Northwestern-UCLA (N-UCLA) Multiview Action 3D } dataset \cite{Wang_2014_CVPR}  consists of 10 classes of actions where every action is acted by 10 subjects. Three Kinect cameras record the action simultaneously and yield 1494 action videos in total.   We adopt the same evaluation setting as in \cite{zhang2020eleatt} by using samples from the first two views for training and the other for testing.
(2) \textbf{NTU RGB+D 60} (NTU 60) dataset \cite{shahroudy2016ntu} is  popular for skeleton based action recognition due to its variety of actions (60 classes) and its large scale (56578 samples). 
We follow the provided evaluation protocol: (a) Cross-Subject (C-Sub) setting  that separates 40091 samples into training set and the rest for testing set by different persons. (b) Cross-View  (C-View) setting  that covers 37646 samples captured by one camera for training and samples from the other camera are for testing. 
(3)  \textbf{NTU RGB+D 120} (NTU 120) dataset \cite{liu2019ntu} is NTU 60 based extension, whose scale is up to 120 classes of actions, 106 participants,  and 113945 sequences in total. Similar as NTU 60, two validation protocols should be followed: (a) Cross-Subject (C-Sub) and (b) Cross-Setup (C-Set). In C-Sub, 63026 samples performed by 53 persons are for training and the others are for testing. In C-Set, all 32 setups  are separated as a half for training and the other half for testing.

\subsection{Configuration Details}

\begin{figure}[]
    \centering
    
        \subfigure[Raw Skeletons]{\scalebox{0.24}{
    \includegraphics{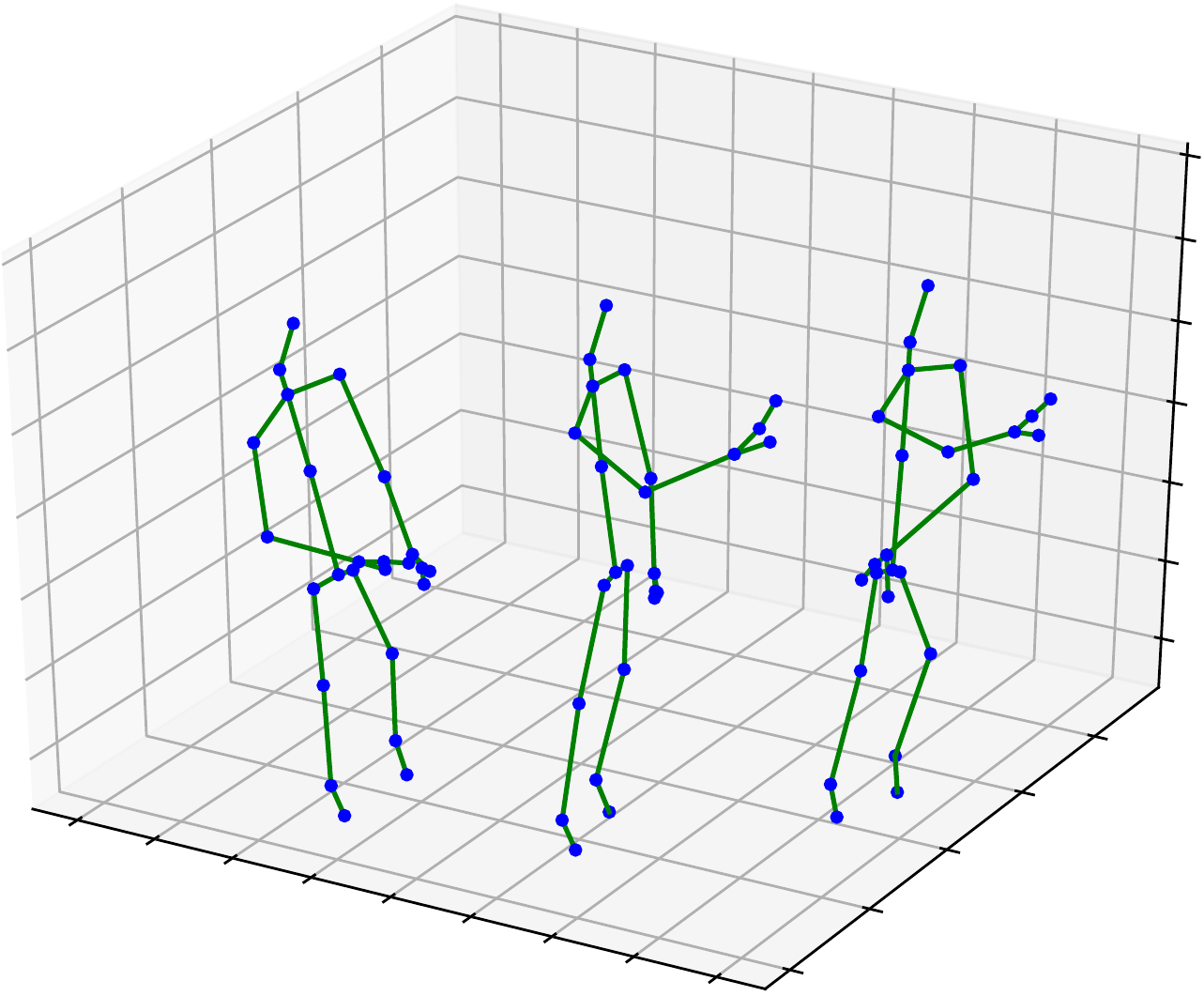}}
    \label{fig:raw}}
    \quad
            \subfigure[View Invariant Skeletons]{\scalebox{0.24}{
    \includegraphics{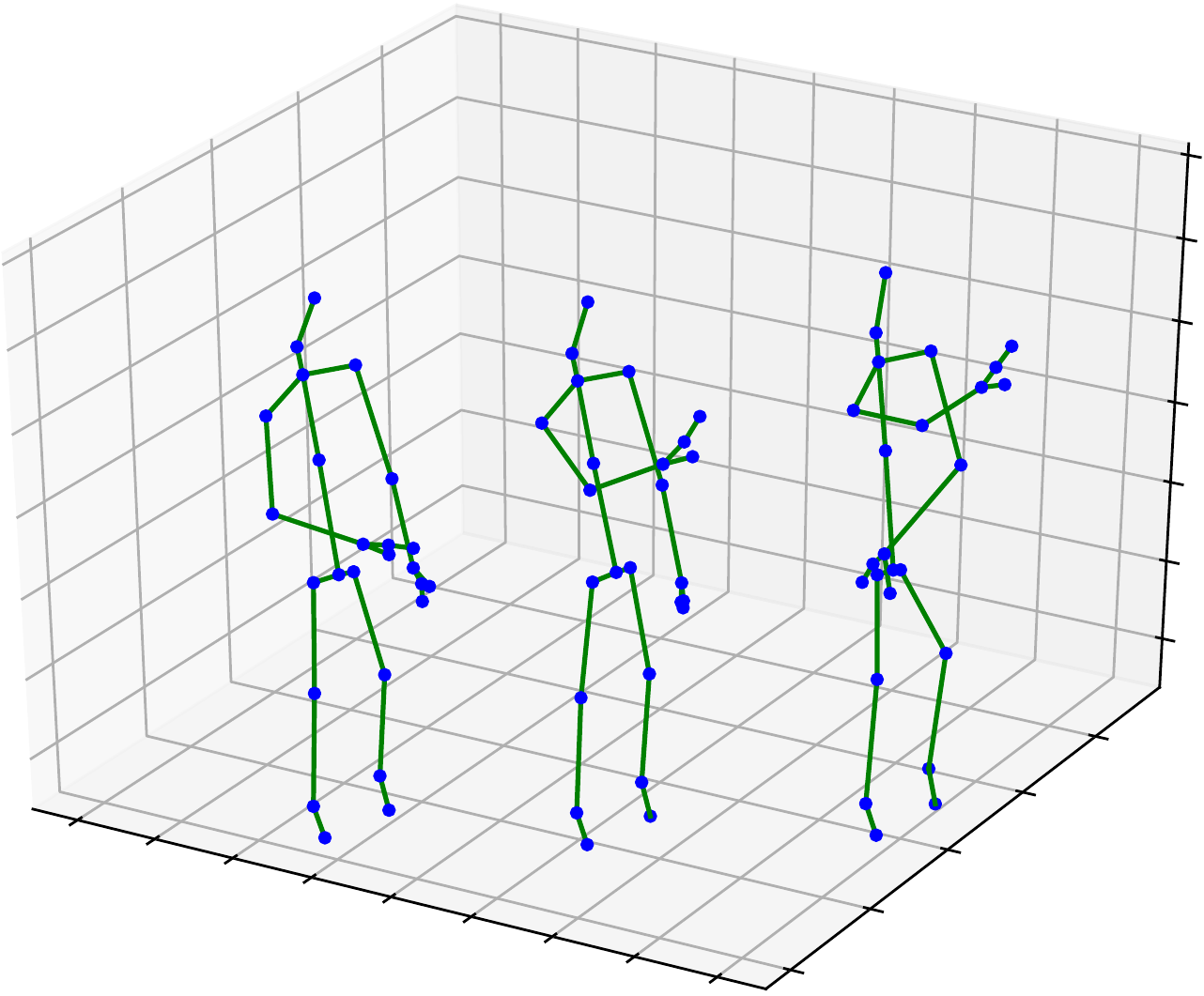}}
    \label{fig:new coord}}
    \quad
           
    % \label{fig:view invariant}
    \caption{}
\end{figure}
\textbf{Pre-processing}: 
To overcome the orientation misalignment of skeleton movements shown in Fig. \ref{fig:raw}, we transform the raw data into a view-invariant coordinate system \cite{lee2017ensemble}  as illustraed in Fig. \ref{fig:new coord}.
The transformed joint coordinates  are then given by: 
\begin{equation}
\boldsymbol{x}_{t,j}=\mathbf{R}^{-1}\left(\boldsymbol{x}_{t,j}-\boldsymbol{o}_{R}\right), \forall j \in J, \, \forall t \in T
\end{equation}
where $\boldsymbol{x}_{t,j} \in \mathbb{R}^{3 \times 1}$. The rotation $\mathbf{R}$ and the origin of rotation $\boldsymbol{o}_R$ are determined by:
\begin{equation}
\mathbf{R}=\left[\frac{\boldsymbol{u}_{1}}{\left\|\boldsymbol{u}_{1}\right\|}\left|\frac{\hat{\boldsymbol{u}}_{2}}{\left\|\hat{\boldsymbol{u}}_{2}\right\|}\right| \frac{\boldsymbol{u}_{1} \times \hat{\boldsymbol{u}}_{2}}{\left\|\boldsymbol{u}_{1} \times \hat{\boldsymbol{u}}_{2}\right\|}\right], \boldsymbol{o}_{R}=x_{1, \text{root}},
\end{equation}
where $\boldsymbol{u}_{1}=x_{1, \text{spine}}-x_{1,\text{root}}$ denotes the vector vertical to the floor and $\hat{\boldsymbol{u}}_{2}=\frac{\boldsymbol{u}_{2}-\text{Proj}_{\boldsymbol{u}_{1}}\left(\boldsymbol{u}_{2}\right)}{\left\|\boldsymbol{u}_{2}-\text{Proj}_{\boldsymbol{u}_{1}}\left(\boldsymbol{u}_{2}\right)\right\|}$
where $\boldsymbol{u}_{2}=x_{1, \text{hip left}}-x_{1,\text{hip right}}$ denotes the difference vector between the left and right hip joints at the initial time step of each sample. $\text{Proj}_{\boldsymbol{u}_{1}}\left(\boldsymbol{u}_{2}\right)$ represents the vector projection of $\boldsymbol{u}_2$ onto $\boldsymbol{u}_1$. 
$\times$ is the cross product and $x_{1,\text{root}}$ is the spine base joint at the initial frame. The sequence length is fixed at 50  and we pad zeros if the sample is less than the fixed length.

PCRP is based on the encoder-decoder structure of  \cite{su2020predict} with fixed weights for the decoder, but we replace Bi-GRU stated in \cite{su2020predict} with  the  Uni-GRU for the encoder. We pre-train PCRP for 50 epochs on the N-UCLA dataset and for  10 epochs on the NTU 60/120 dataset. The learning rate is 0.001 in pre-training stage. In the linear evaluation, we fix the encoder and train the linear classifier by 50 epochs on the N-UCAL dataset and by 30 epochs on the NTU 60/120 dataset.  The learning rate is 0.01 in evaluation stage. Adam  is applied for model optimization.

\subsection{Performance Comparison}
We compare our  PCRP with previous relevant  unsupervised learning methods, supervised methods, and hand-crafted methods on three large datasets including N-UCLA dataset, NTU 60 dataset, and NTU 120 dataset. The performance comparisons are shown in  Table \ref{tab:ucla}, \ref{tab: ntu60}, \ref{tab:ntu120}. For  an  unsupervised learning method P\&C FW \cite{su2020predict}, we implement it  on linear evaluation protocol instead of  KNN evaluation, and also 
rid the auto-encoder part   to be efficient in pre-training but not compromising much the performance.

% ========== ucla sota
\begin{table}[t]
\centering
\caption{Comparison with prior methods on N-UCLA dataset. ``*'' represents depth image based methods. Bold numbers refer to the best performers.}
\scalebox{0.7}{
\begin{tabular}{@{}lll@{}}
\toprule
\textbf{Id} & \textbf{Method}      & \textbf{Acc (\%)} \\ \midrule
& \textbf{Hand-Crafted Methods} &   \\ \midrule
    \textbf{1}        & Lie Group   \cite{vemulapalli2014human}         & 74.2                 \\
    \textbf{2}        & Actionlet Ens \cite{wang2013learning}       & 76.0                 \\ \midrule
 & \textbf{Supervised Methods}   &                      \\ \midrule
    \textbf{3}        & HBRNN-L \cite{du2015hierarchical}             & 78.5                 \\ \midrule
            & \textbf{Unsupervised Methods} &                      \\ \midrule
    \textbf{4}    & AS-CAL\cite{rao2020augmented} & 35.6\\    
    \textbf{5}        & *Luo \textit{et al.}  \cite{luo2017unsupervised}         & 50.7                 \\
\textbf{6} & *Li \textit{et al.}  \cite{li2018unsupervised}          & 62.5                 \\
\textbf{7} & LongT GAN\cite{zheng2018unsupervised}            & 74.3                 \\
\textbf{8}& MS$^2$L \cite{lin2020ms2l} & 76.8 \\
\textbf{9}& P\&C FW\cite{su2020predict} & 83.3 \\
\textbf{10}& PCRP (Ours)                 &      \textbf{87.0}                \\ \bottomrule
\end{tabular}}
\label{tab:ucla}
\end{table}

\subsubsection{Comparison with Unsupervised Methods}
As shown in Table \ref{tab:ucla} on N-UCLA dataset, the proposed PCRP shows 3.7-24.5\% margin over the state-of-the-art unsupervised methods (Id = 6, 7, 8, 9), which are also based on the encoder-decoder structure to learn action representation. Although  they possess cross-view decoding \cite{li2018unsupervised}, additional adversarial training strategies \cite{zheng2018unsupervised},  decoder-weakening mechanism \cite{su2020predict} or multi-task learning \cite{lin2020ms2l},  they just aim at plain sequential prediction in order and do not consider high-level semantic information learning. 
In contrast, the proposed PCRP is able to simultaneously learn semantic similarity between sequences and enhance action representation learning via reverse prediction. In particular,  our method achieves over 10\% improvement than   Li \textit{et al.} (Id = 6) that focus on view-invariant action
representation learning, which validates the superior robustness of our method to viewpoint variations. 
On the other hand, our approach takes skeleton sequences as inputs that are smaller sized than depth images,  but it still   significantly outperforms depth-image based  methods (Id = 5, 6). 
Above advantages   of our approach  are also similarly shown on NTU 60 dataset (see Table \ref{tab: ntu60}) and NTU 120 dataset (see Table \ref{tab:ntu120}). 
These comparison results do show  the effectiveness and efficacy of the proposed PCRP. 

% =========== ntu 60 sota
\begin{table}[t]
\centering
\caption{Comparison with prior methods on NTU  60 dataset.  Bold numbers refer to the best unsupervised performers. }
% 4.0
\setlength{\tabcolsep}{3.1mm}{
\scalebox{0.7}{
\begin{tabular}{@{}llcc@{}}
\toprule
    %   &    & \multicolumn{2}{c}{\quad \textbf{NTU RGB+D 60}} \\
&
& \textbf{C-View} &  \textbf{C-Sub}  \\
\textbf{Id} & \textbf{Method}  &\textbf{Acc (\%)} &
   \textbf{Acc (\%)} \\ \midrule
 &\textbf{Hand-Crafted Methods}  &&\\  \midrule
\textbf{1} &*HON4D \cite{Ohn-Bar_2013_CVPR_Workshops}  & 7.3& 30.6\\
\textbf{2} &*Super Normal Vector \cite{Yang_2014_CVPR} & 13.6 & 31.8\\
\textbf{3} &*HOG$^2$\cite{Oreifej_2013_CVPR} & 22.3 & 32.2\\
\textbf{4} &Skeletal Quads \cite{evangelidis2014skeletal}   & 41.4& 38.6\\ 
\textbf{5} &Lie Group \cite{vemulapalli2014human}                      &        52.8  &      50.1   \\ 

\midrule
&\textbf{Supervised Methods} && 
                     \\ \midrule
% \textbf{6}  & RNN & 23.4  & 18.9\\
\textbf{6} & HBRNN\cite{du2015hierarchical}                      &    64.0   &   59.1  \\  
\textbf{7} & Deep RNN \cite{shahroudy2016ntu}  & 64.1 & 56.3 
\\ 
% \textbf{9}  & GRU  & 72.8 & 70.2\\
% \textbf{10}  & LSTM & 72.7  & 53.9 \\

\midrule
&\textbf{Unsupervised Methods}     & \multicolumn{1}{l}{} & \multicolumn{1}{l}{}                     \\ \midrule

\textbf{8} & *Shuffle\&Learn\cite{misra2016shuffle}         &      40.9     &     46.2 \\
\textbf{9} & *Luo \textit{et al.}\cite{luo2017unsupervised}       &      53.2      &       61.4      \\
\textbf{10} &*Li \textit{et al.}\cite{li2018unsupervised}             &  53.9        &   60.8  \\
\textbf{11} &LongT GAN\cite{zheng2018unsupervised}         &   48.1          &   39.1   \\
\textbf{12} & P\&C FW \cite{su2020predict}  & 44.3 & 50.8\\
\textbf{13} & MS$^2$L  \cite{lin2020ms2l} & - & 52.6 \\
% \textbf{13} & AS-CAL\cite{rao2020augmented} & 63.6   &  \textbf{58.0} \\
\textbf{14} & PCRP (Ours) & \textbf{63.5}& \textbf{53.9}\\
 \bottomrule

\end{tabular}
}
\label{tab: ntu60}}
\end{table}

% =============== ntu 120 sota
\begin{table}[t]
\centering
\caption{Comparison with supervised and unsupervised  methods on NTU  120 dataset. Bold numbers denote the best performers.}
% 3.8
\setlength{\tabcolsep}{3.5mm}{
\scalebox{0.7}{
\begin{tabular}{@{}llcc@{}}
\toprule
    %  &       & \multicolumn{2}{c}{\quad \textbf{NTU RGB+D 120}} \\
&& \textbf{C-Set} &
  \textbf{C-Sub} \\
\textbf{Id}&\textbf{Method}  &\textbf{Acc (\%)} &
   \textbf{Acc (\%)} \\\midrule
% &\textbf{Hand-crafted Methods}  &&\\   \midrule
% \textbf{1}&Dynamic Skeleton\cite{hu2015jointly}              &   54.7  &  50.8                \\ \midrule
&\textbf{Supervised  Methods}     & \multicolumn{1}{l}{} & \multicolumn{1}{l}{} \\ \midrule
% \textbf{2}  & RNN & 6.3  & 7.2 \\
\textbf{1}&Part-Aware LSTM\cite{shahroudy2016ntu} & 26.3    &  25.5      \\
\textbf{2}&Soft RNN\cite{hu2018early}                       &  44.9          & 36.3          \\
% \textbf{4}  & GRU &  61.6 & 61.2 \\
% \textbf{5}  & LSTM & 58.1 & 41.2 \\
          
 \midrule
&\textbf{Unsupervised  Methods}   & \multicolumn{1}{l}{} & \multicolumn{1}{l}{} \\ \midrule
\textbf{3}& P\&C FW\cite{su2020predict}  & 42.7  & 41.7\\
% \textbf{4}& AS-CAL\cite{rao2020augmented} & 49.2 & 48.3 \\
\textbf{4} &PCRP (Ours)   & \textbf{45.1}  &  \textbf{41.7}  \\
 \bottomrule
\end{tabular}}
}
\label{tab:ntu120}
\end{table}

% =================== loss & acc curve
\begin{figure*}[t]
    \centering
    % 0.52
    \subfigure[N-UCLA]{    \scalebox{0.28}{
    \includegraphics{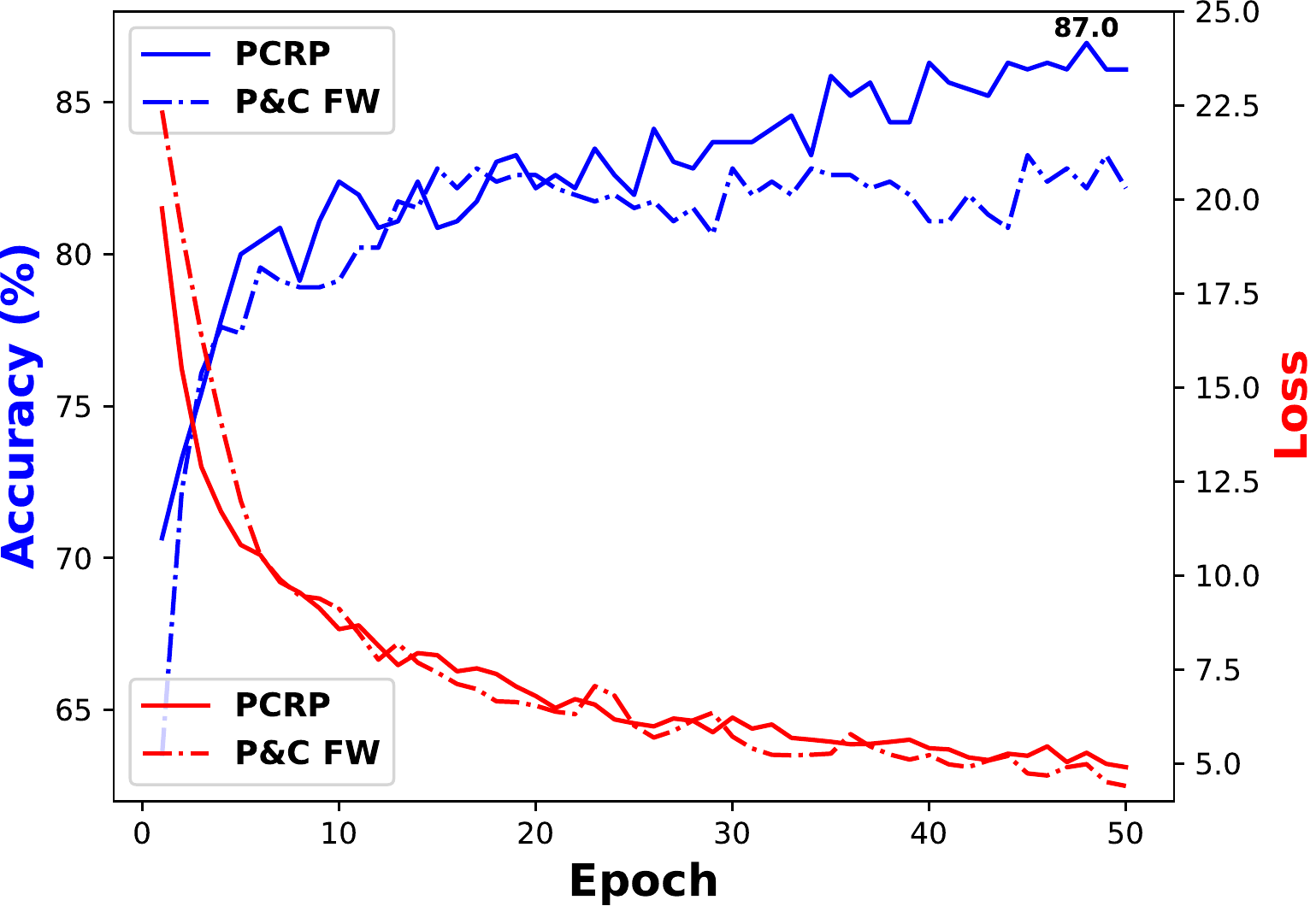}    }
    \label{fig:ucla trainloss}}
    \subfigure[NTU 60 C-Sub]
    {\scalebox{0.28}{\includegraphics{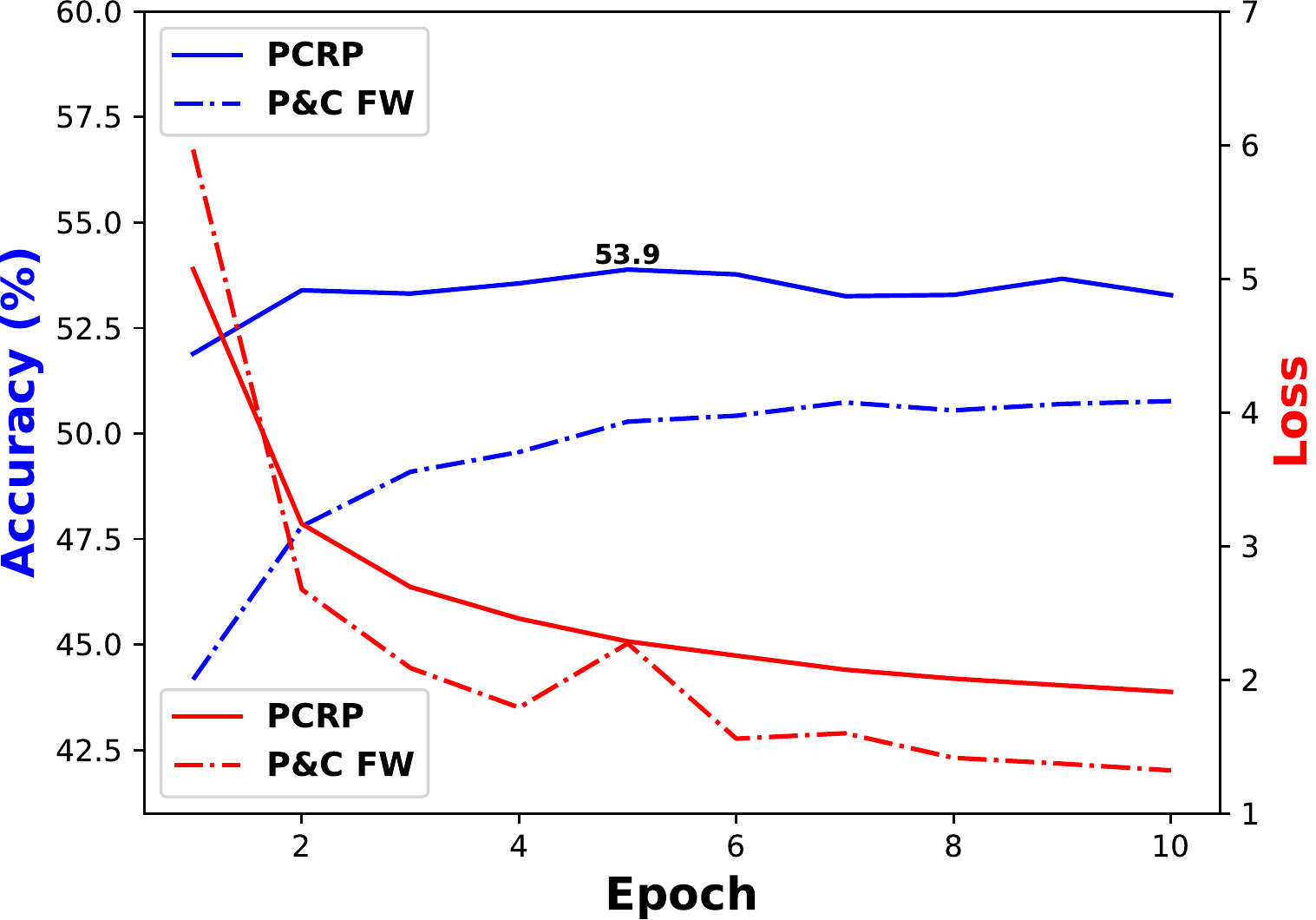}} \label{fig:ntu60sub trainloss}}
\quad
    \subfigure[NTU 60 C-View]
    {\scalebox{0.28}{\includegraphics{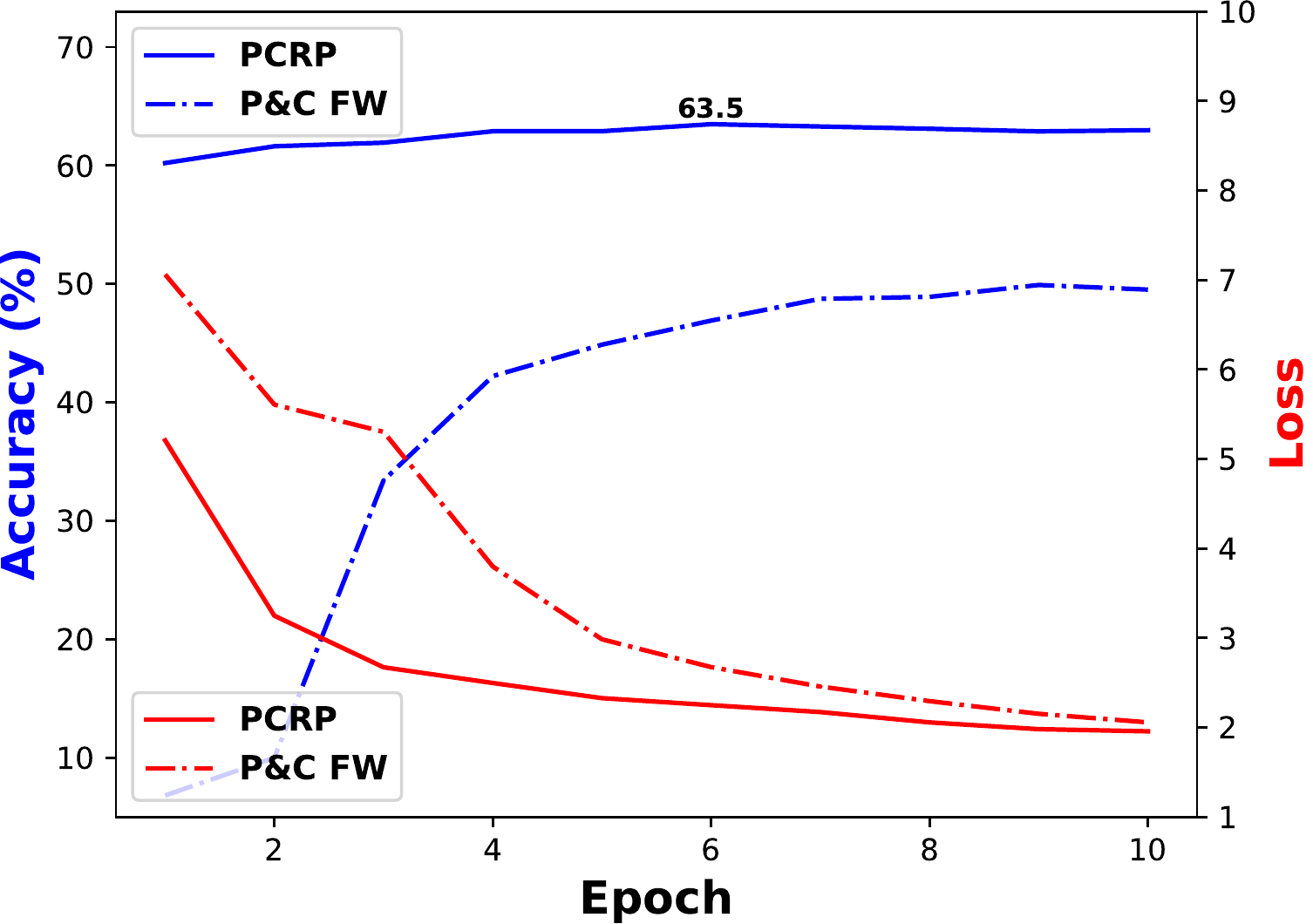}} \label{fig:ntu60view trainloss}}
    \quad
      \subfigure[NTU 120 C-Sub]
      {\scalebox{0.28}{\includegraphics{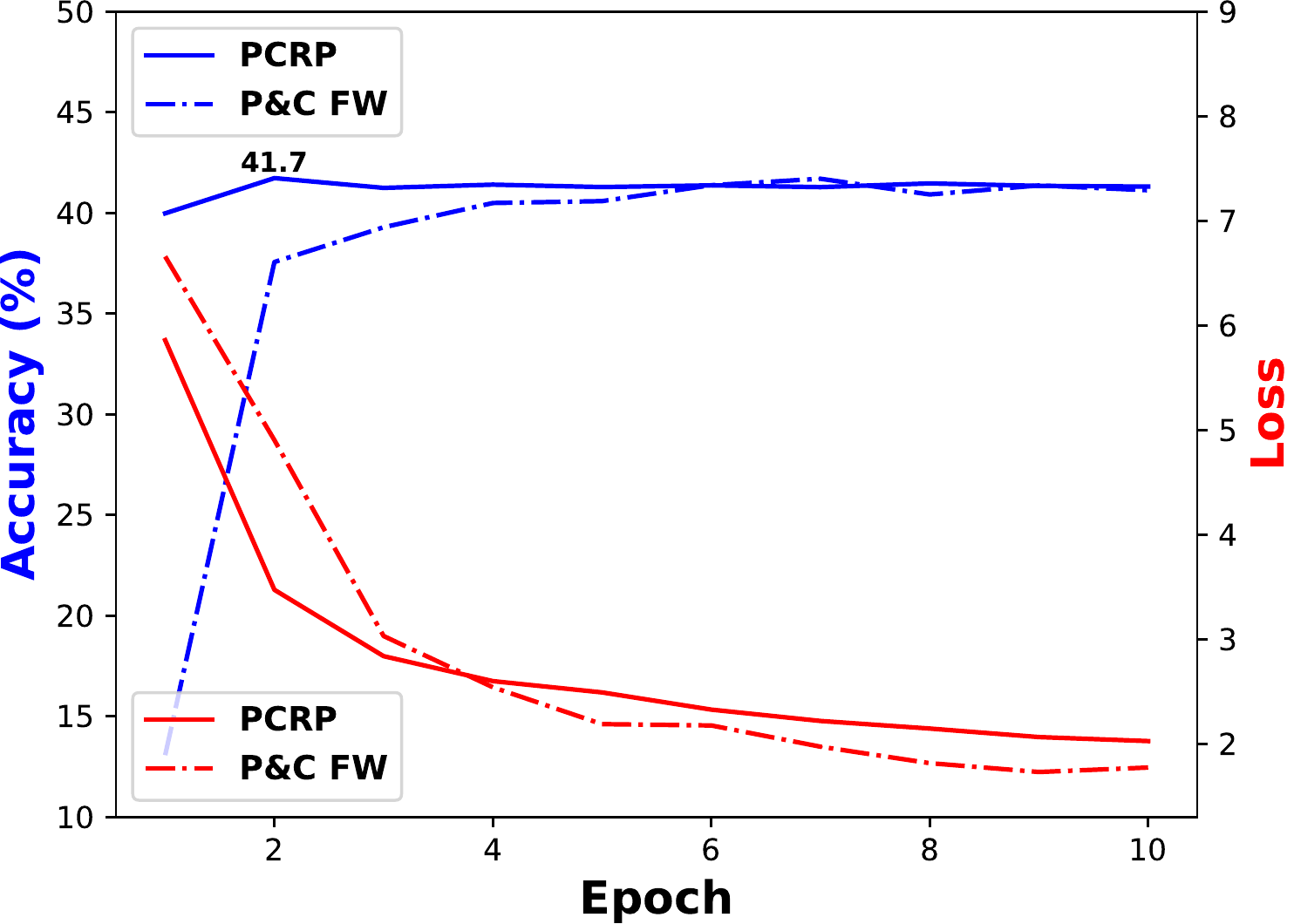}} \label{fig:ntu120sub trainloss}}
    \quad
     \subfigure[NTU 120 C-Set]
     {\scalebox{0.28}{\includegraphics{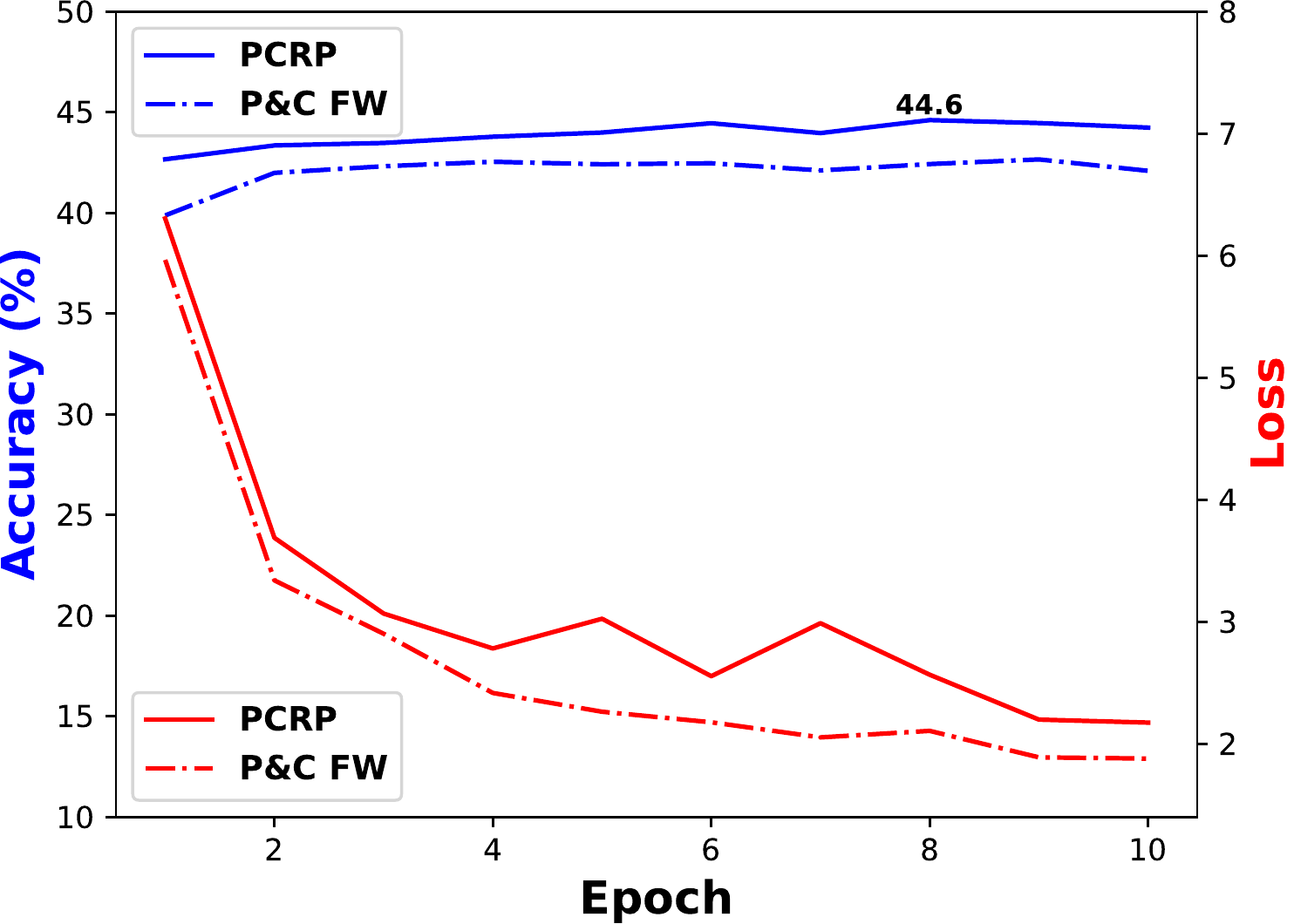}}\label{fig:ntu120view trainloss}}

    \caption{Pre-training loss curves (red) and linear evaluation accuracy curves (blue) for three datasets. 
     } 
    \label{fig:zhexiantu}
\end{figure*}

\begin{figure}[h]
    \centering
        \subfigure[N-UCLA]{\scalebox{0.23}{
    \includegraphics{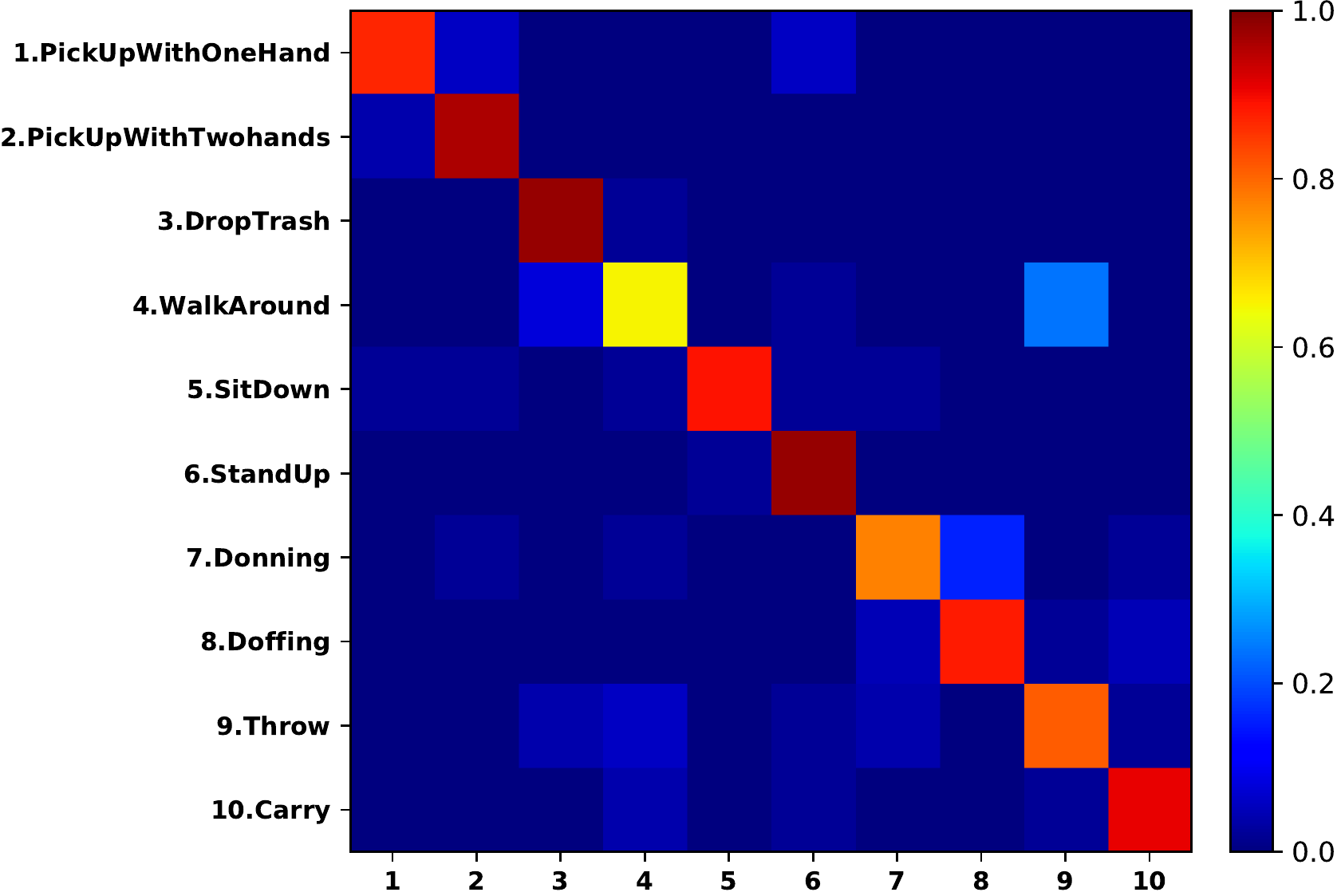}}}
    \,
            \subfigure[NTU 60 C-View]{\scalebox{0.23}{
    \includegraphics{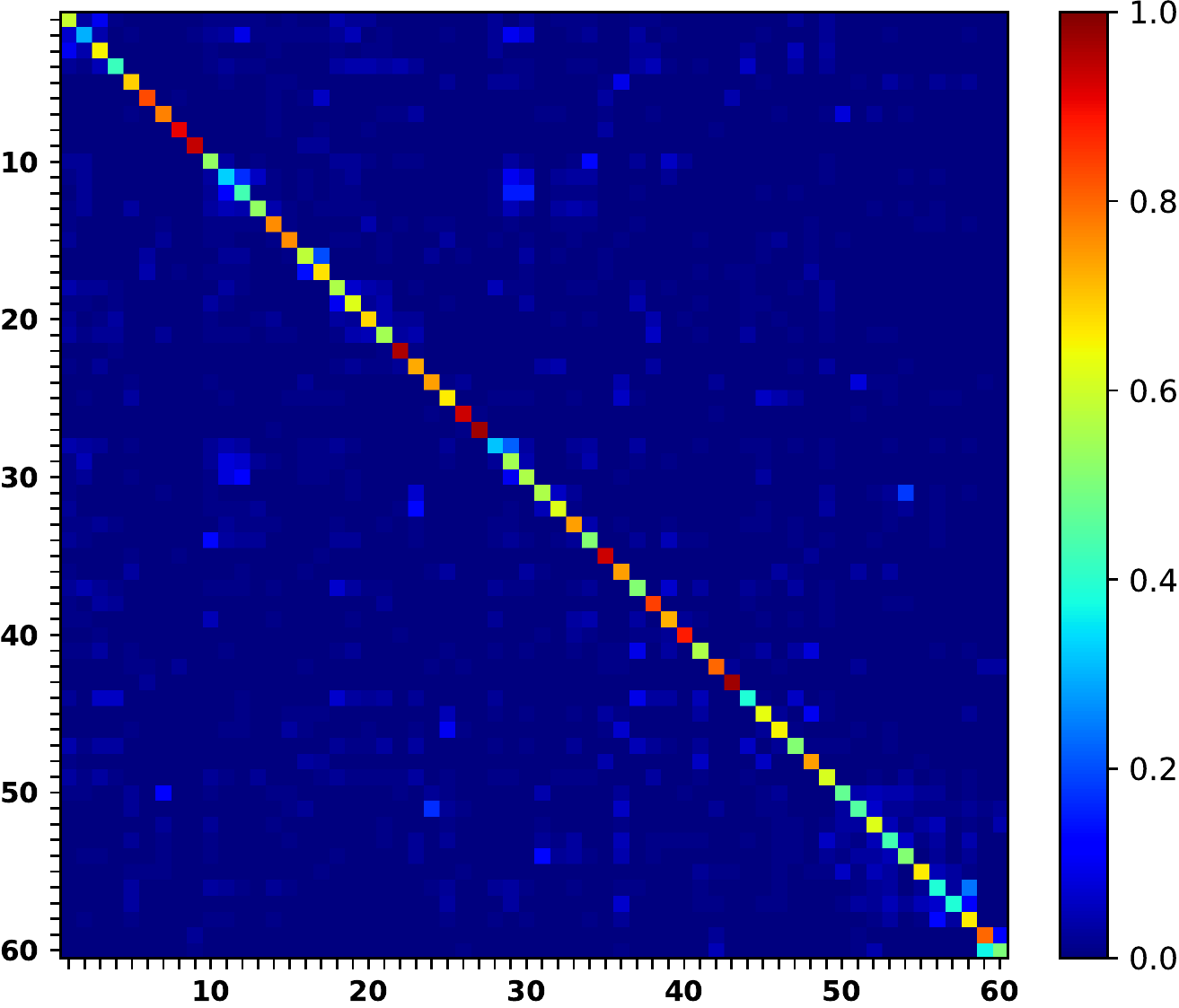}}}
    \,
    \subfigure[NTU 120 C-Set]{\scalebox{0.23}{
    \includegraphics{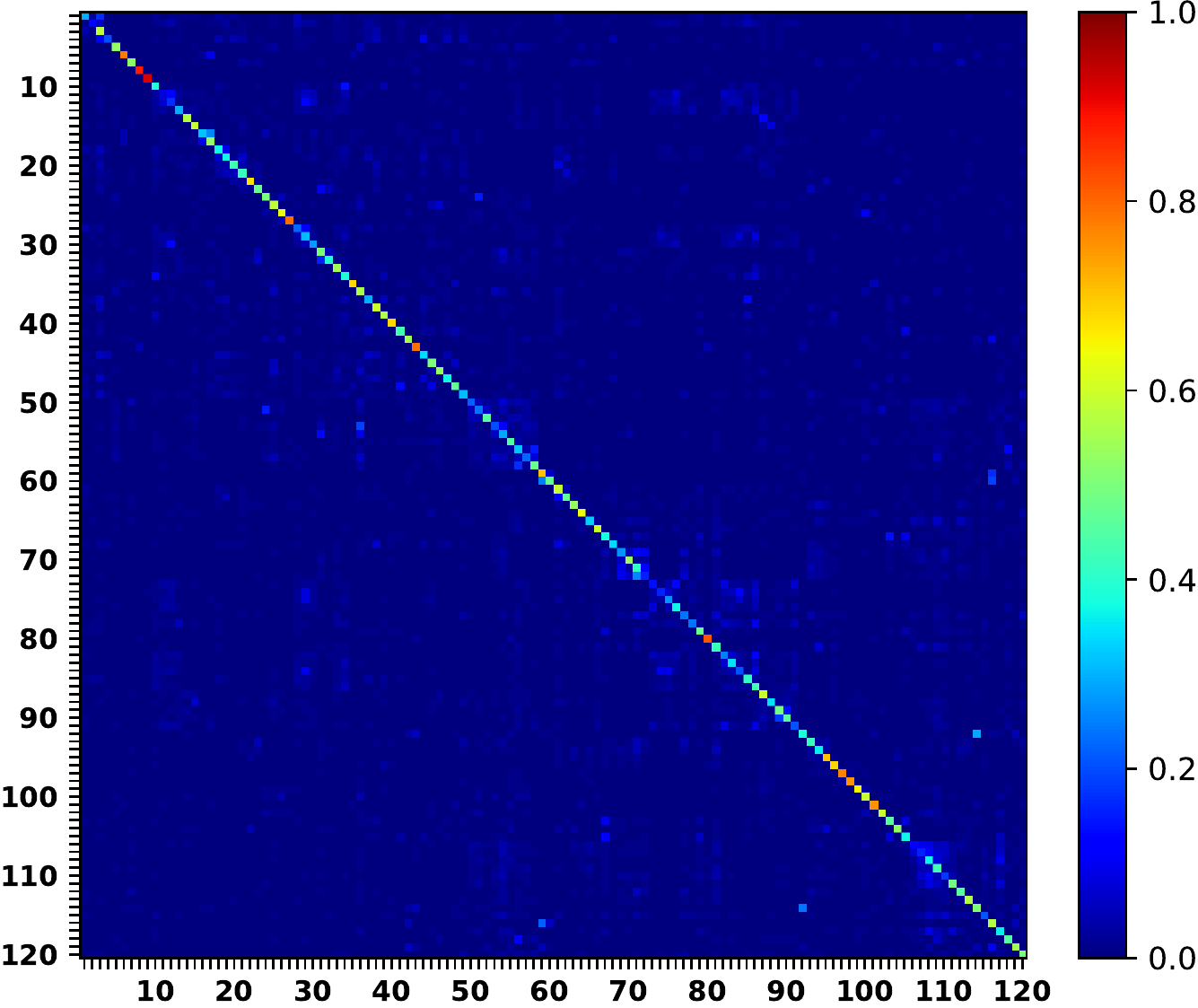}}}
    \,
    \caption{Confusion Matrix}  
    \label{fig:confusion matrix}
\end{figure}

Since our work is based on P\&C FW \cite{su2020predict}, we make further comparison  of our PCRP with  P\&C FW on pre-training loss curves and  evaluation accuracy curves.  In Fig. \ref{fig:ucla trainloss} on N-UCLA dataset, we observe that PCRP shows increasing margin than 
P\&C FW as epoch goes on. 
When it comes to  larger scale datasets, \textit{i.e.,} NTU 60/120 dataset (see Fig. \ref{fig:ntu60sub trainloss}-\ref{fig:ntu120view trainloss}), the proposed work shows  great superior over  P\&C FW that PCRP keeps  high evaluation accuracy from the beginning  while P\&C FW's accuracy grows increasingly. We here argue that excellent unsupervised learning methods should be of high efficiency  that they do not require too many pre-training epochs to achieve  high evaluation accuracy, and they are supposed to  maintain it as the epoch increases. From this point, our method indeed performs better than  P\&C FW. We plot  confusion matrix results in Fig. \ref{fig:confusion matrix} 
% ============ confusion matrix

\subsubsection{Comparison with Hand-Crafted and Supervised Methods}
The proposed PCRP significantly surpasses several  hand-crafted methods (Id = 1-2 in Table \ref{tab:ucla}) on the N-UCLA dataset and  (Id = 1-5 in Table \ref{tab: ntu60}) on the  NTU 60 dataset. For instance in Table  \ref{tab: ntu60}, PCRP shows better results than Skeletal Quads (Id = 4) and Lie group (Id = 5) by at least 10.7\% on the  C-View protocol and 3.8\% on the  C-Sub protocol. In addition, the proposed work is   competitive or superior  compared with some  prior supervised methods on three datasets: (1) For the N-UCLA dataset in Table \ref{tab:ucla}, the proposed work has 6.5\% margin better than HBRNN (Id = 3). (2) For the NTU 60 dataset in Table \ref{tab: ntu60}, our method shows comparable results with Deep RNN  (Id = 7).  (3) For the NTU 120 dataset in Table \ref{tab:ntu120}, our proposed approach presents advantage over Part-Aware LSTM (Id = 1)  and Soft RNN (Id = 2) by 0.2-18.8\% on different evaluation settings.

\subsection{Ablation Study}
In this section, we conduct extensive ablation experiments on three datasets mentioned above to provide solid validation of our proposed work. 

\begin{figure}[]
    \centering
  
        \subfigure[NTU 60 sub]{\scalebox{0.28}{
    \includegraphics{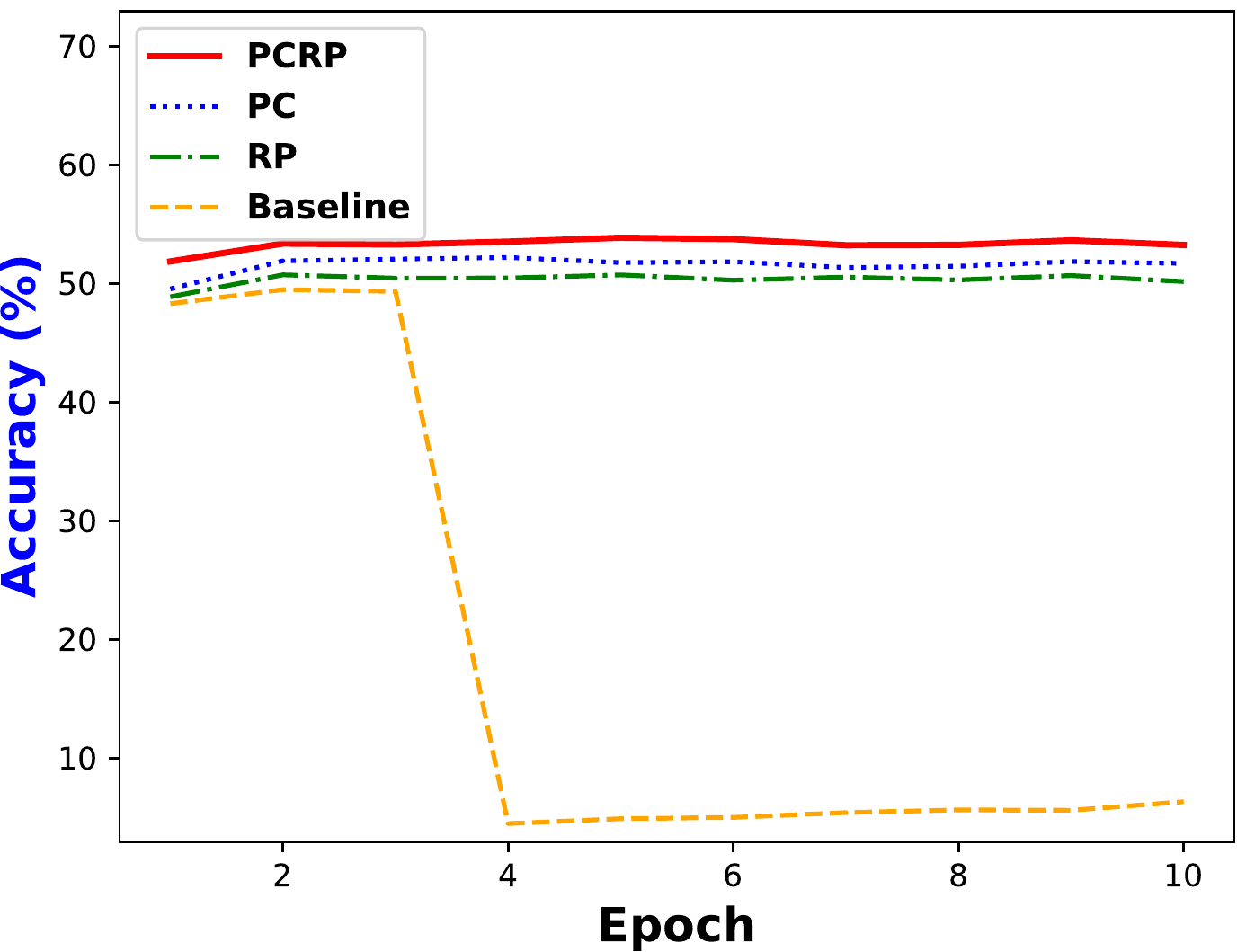}}\label{fig:60sub 4curves}}
    \,
            \subfigure[NTU 60 C-View]{\scalebox{0.28}{
    \includegraphics{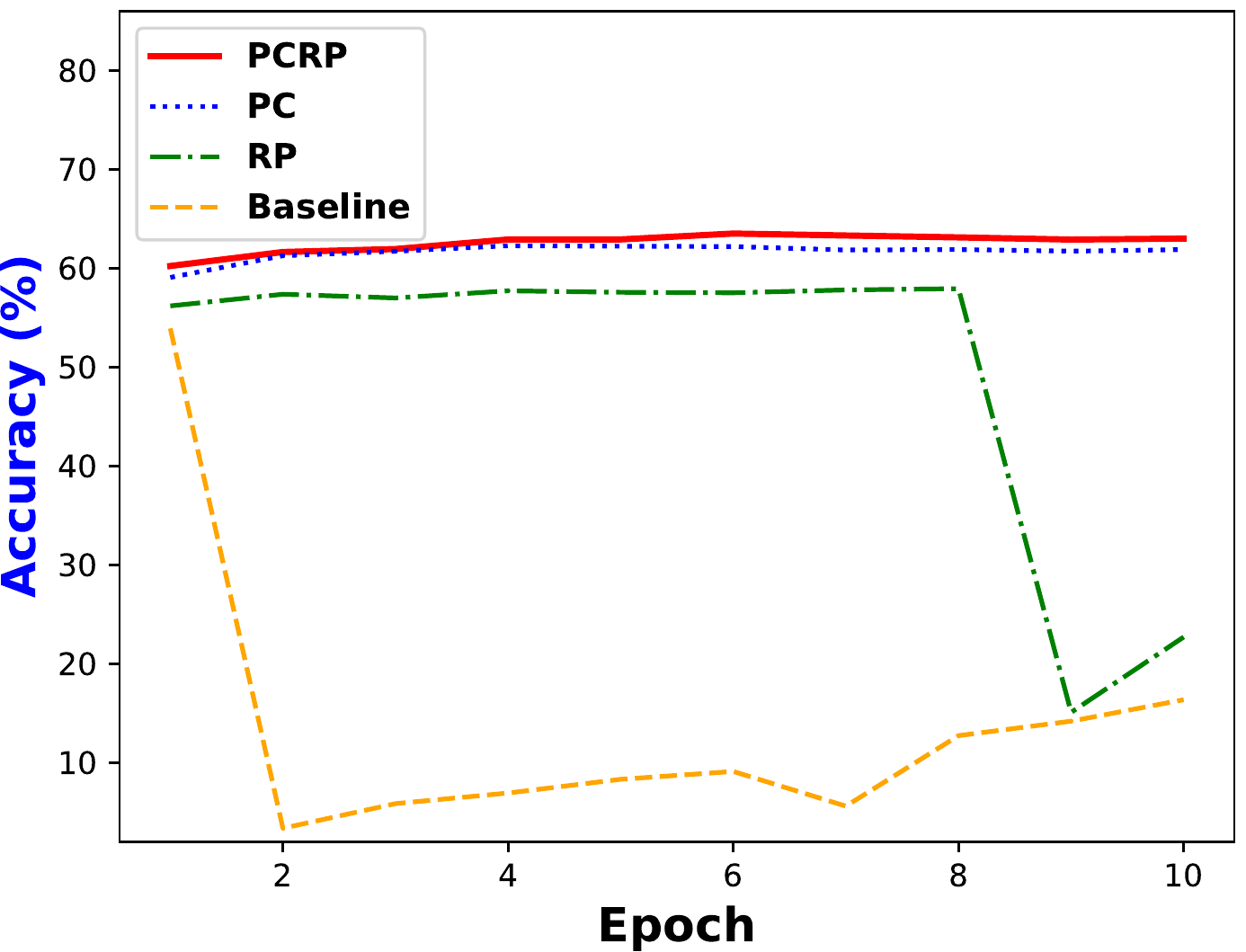}}\label{fig:60view 4curves}}
    \,
    \subfigure[NTU 120 C-Sub]{\scalebox{0.28}{
    \includegraphics{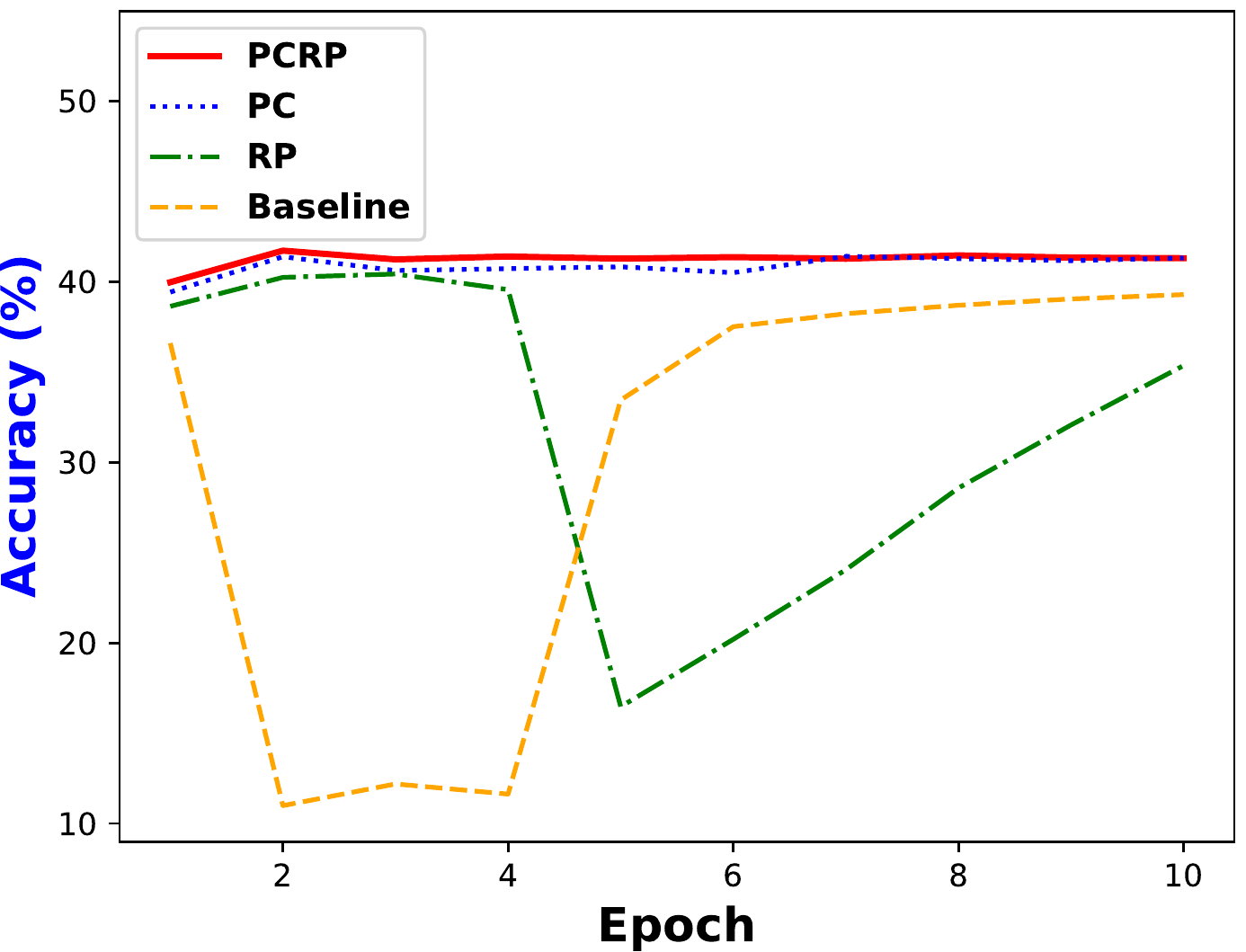}}\label{fig:120sub 4curves}}
    \,
        \subfigure[NTU 120 C-Set]{\scalebox{0.28}{
    \includegraphics{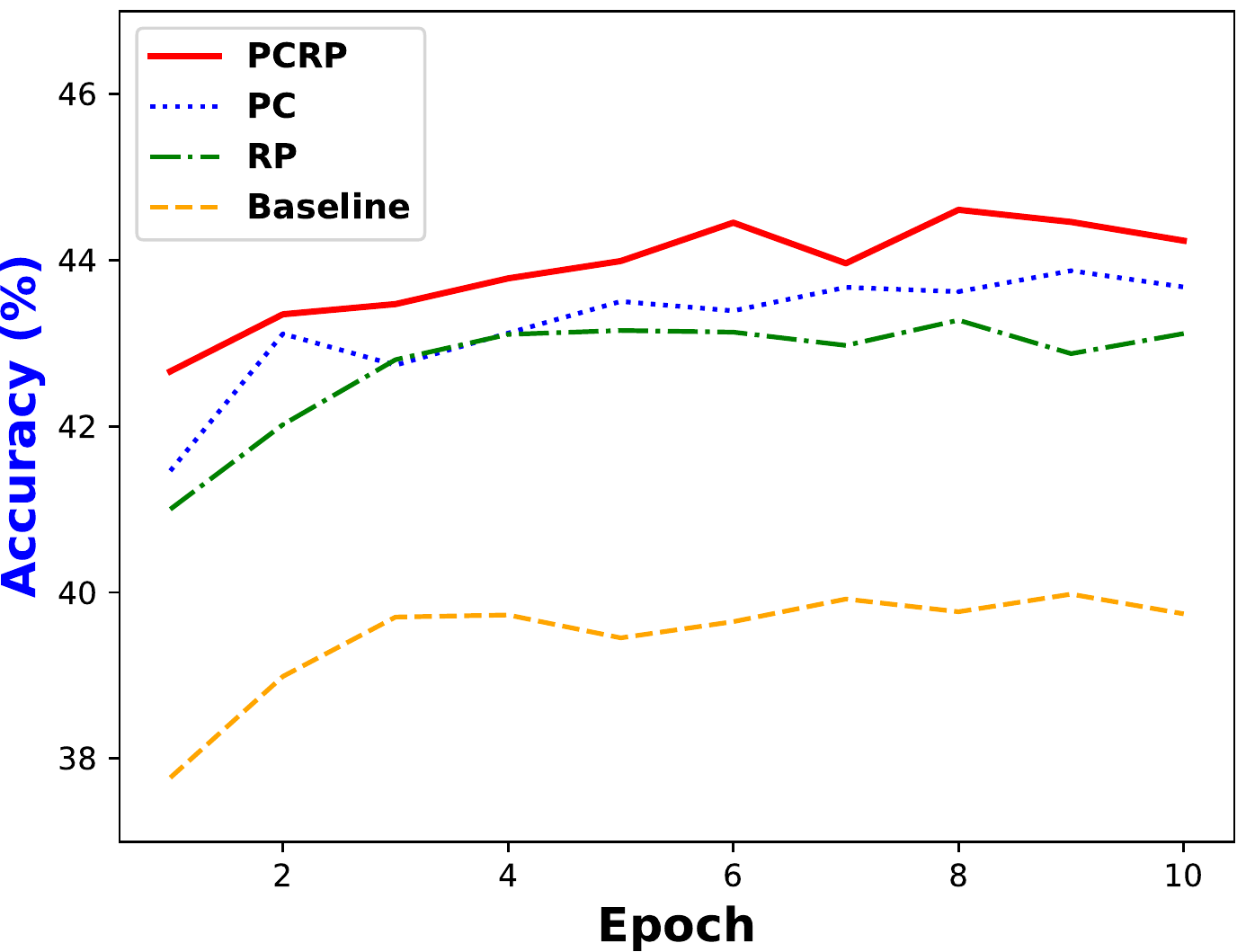}}\label{fig:120view 4curves}}
    \caption{Linear evaluation accuracy curves of PCRP, PC, RP, and baseline on NTU 60/120 dataset.}  
  
\end{figure}

\subsubsection{Analysis of  PC and RP}
In this part, we explore the role of prototypical contrast (PC) and reverse prediction (RP). The baseline is P\&C FW \cite{su2020predict} with Uni-GRU encoder instead of Bi-GRU  stated in \cite{su2020predict}. When the experiment is involved in PC, we run $M=3$  times clustering with different cluster number  (see Eq. \ref{eq:protomae}). 

In  Table \ref{tab:abla ucla} for the N-UCLA dataset, compared with the baseline (Id = 1), RP (Id = 2) presents 0.9\% improvement, which validates the effectiveness of RP in our framework.  This effectiveness can also be observed from the comparison between (Id = 3) and (Id = 4). For the effective function of PC, the item (Id = 3) runs 3 times clustering with 40, 70, 100 clusters respectively and it shows superior performance over the baseline (Id = 1) by 2.5\%. Besides, the item (Id = 4) also shows 2.8\% margin higher than (Id = 2). Combing PC and RP, the final model (Id = 4) achieves the best result. In the  larger datasets such as NTU 60/120 dataset, the effectiveness of PC and RP can also be demonstrated and shown in Table \ref{tab:abla ntu60} and Table \ref{tab:abla ntu120}.

Furthermore, we plot  evaluation accuracy curves of PCRP, PC, RP, and baseline  on NTU 60/120 dataset. As shown in Table \ref{fig:60sub 4curves}-\ref{fig:120view 4curves}, our  approach PCRP (red line) is able to obtain high evaluation accuracy at beginning and then maintain it as the pre-training goes on, which shows its powerful and robust action representation learning.

\subsubsection{Effects of Various Number of Clusters}
To understand the effects of  $M$ times running with  different cluster number, we conduct ablation experiments on the N-UCLA dataset, the NTU 60 dataset in C-View setting, and the NTU 120 dataset in C-Set setting. As shown in Table \ref{tab:cluster of ucla}, $M = 1$ with 70 clusters for PC (Id = 2) obtains 87.0\%, which outperforms $M = 3$ times running for PC (Id = 4-5). Likewise, Table \ref{tab:cluster of ntu120 set} has similar observation. In Table \ref{tab:cluster of ntu60 view}, $M = 3$ with 90, 120, 150 (Id = 5) achieves 63.5\%, the highest score among (Id = 1-6). Along this line, we can understand running larger  $M$ with different  cluster number does not necessarily guarantee better representation.

% \begin{table}[]
% \centering
% \caption{template}
% \scalebox{0.7}{
% }

% \end{table}

% ================= ablation 
\begin{table}[]
\centering
\caption{Ablation Experiments of PCRP.}
\subtable[N-UCLA]{\scalebox{0.7}{
\setlength{\tabcolsep}{0.9mm}{
\begin{tabular}{@{}llcl@{}}
\toprule
\textbf{Id} & \textbf{RP} & \multicolumn{1}{l}{\textbf{Clusters}} & \textbf{Acc (\%)} \\ \midrule
\textbf{1}  & \XSolid           & \XSolid                                          & 82.6              \\
\textbf{2}  & \checkmark         & \XSolid                                          & 83.5              \\
\textbf{3}  & \XSolid           & 40,70,100                             & 86.1              \\
\textbf{4}  & \checkmark         & 40,70,100                             & \textbf{86.3}               \\ \bottomrule
\end{tabular}
}\label{tab:abla ucla}}

}
\quad
\subtable[NTU 60]{\scalebox{0.7}{
\begin{tabular}{@{}llcll@{}}
\toprule
            &             & \multicolumn{1}{l}{}                  & \textbf{C-View}   & \textbf{C-Sub}    \\
\textbf{Id} & \textbf{RP} & \multicolumn{1}{l}{\textbf{Clusters}} & \textbf{Acc (\%)} & \textbf{Acc (\%)} \\ \midrule
\textbf{1}  & \XSolid           & \XSolid                                          & 53.9              & 49.5              \\
\textbf{2}  & \checkmark         & \XSolid                                          & 57.9              & 50.8              \\
\textbf{3}  & \XSolid           & 90,120,150                            & 62.3              & 52.2              \\
\textbf{4}  & \checkmark         & 90,120,150                            & \textbf{63.5}     & \textbf{53.9}     \\ \bottomrule
\end{tabular}
}
\label{tab:abla ntu60}
}
\quad
\subtable[NTU 120]{\scalebox{0.7}{
\begin{tabular}{@{}llcll@{}}
\toprule
            &             & \multicolumn{1}{l}{}                  & \textbf{C-Set}    & \textbf{C-Sub}    \\
\textbf{Id} & \textbf{RP} & \multicolumn{1}{l}{\textbf{Clusters}} & \textbf{Acc (\%)} & \textbf{Acc (\%)} \\ \midrule
\textbf{1}  & \XSolid           & \XSolid                                          & 40.0              & 39.3              \\
\textbf{2}  & \checkmark         & \XSolid                                          & 43.3              & 40.4              \\
\textbf{3}  & \XSolid           & 150,180,210                           &     43.9              & 41.4              \\
\textbf{4}  & \checkmark         & 150,180,210                           & \textbf{44.6}     & \textbf{41.7}     \\ \bottomrule
\end{tabular}
}
\label{tab:abla ntu120}
}

\end{table}

% ============ cluster ablation 
\begin{table}[]
\centering
\caption{Ablation Experiments of $M$ of Eq. \ref{eq:protomae}}
\subtable[N-UCLA]{\scalebox{0.7}{
\setlength{\tabcolsep}{1.2mm}{
\begin{tabular}{lccl}
\toprule
\textbf{Id} & $\boldsymbol{M}$ & \multicolumn{1}{l}{\textbf{Clusters}} & \textbf{Acc (\%)} \\ \midrule
\textbf{1}  & 1 & 40                                    & 86.5              \\
\textbf{2}  & 1 & 70                                    & \textbf{87.0}     \\
\textbf{3}  & 1 & 100                                   & 85.4              \\
\textbf{4}  & 3 & 30,60,90                              & 86.5              \\
\textbf{5}  & 3 & 40,70,100                             & 86.3              \\
\textbf{6}  & 3 & 50,80,110                             & 86.3              \\ \bottomrule
\end{tabular}}
}
\label{tab:cluster of ucla}}
\,
\subtable[NTU 60 C-View]{
\scalebox{0.7}{
\setlength{\tabcolsep}{0.9mm}{
\begin{tabular}{lccl}
\toprule
\textbf{Id} & $\boldsymbol{M}$ & \multicolumn{1}{l}{\textbf{Clusters}} & \textbf{Acc (\%)} \\ \midrule
\textbf{1}  & 1 & 90                                    & 62.5              \\
\textbf{2}  & 1 & 120                                   & 62.8              \\
\textbf{3}  & 1 & 150                                   & 62.9              \\
\textbf{4}  & 3 & 80,110,140                            & 62.6              \\
\textbf{5}  & 3 & 90,120,150                            & \textbf{63.5}     \\
\textbf{6}  & 3 & 100,130,160                           & 62.8              \\ \bottomrule
\end{tabular}}
}
\label{tab:cluster of ntu60 view}}
\,
\subtable[NTU 120 C-Set]{
\scalebox{0.7}{
\setlength{\tabcolsep}{1.2mm}{
\begin{tabular}{@{}lccl@{}}
\toprule
\textbf{Id} & $\boldsymbol{M}$ & \multicolumn{1}{l}{\textbf{Clusters}} & \textbf{Acc (\%)} \\ \midrule
\textbf{1}  & 1 & 150                                   & 43.6              \\
\textbf{2}  & 1 & 180                                   & \textbf{45.1}     \\
\textbf{3}  & 1 & 210                                   & 44.1              \\
\textbf{4}  & 3 & 150,180,210                           & 44.6              \\
\textbf{5}  & 3 & 160,190,220                            & 43.9              \\
\textbf{6}  & 3 & 170,200,230                           & 43.2              \\ \bottomrule
\end{tabular}}
}
\label{tab:cluster of ntu120 set}}

\end{table}

\section{Conclusion}
This paper presents a  novel framework named prototypical contrast and reverse prediction (PCRP) for skeleton-based action recognition. In the view of EM algorithm, PCRP alternately performs  E-step as generating  action prototypes by clustering action encoding from the encoder, and M-step as updating the encoder by contracting the distribution around the prototypes and simultaneously predicting sequences in reverse order. Experiments on three large datasets  show that our  work can learn distinct  action representations and surpasse previous unsupervised approaches.

{\small
\bibliographystyle{ieee_fullname}
\bibliography{ref}
}

\end{document}